\documentclass[letterpaper]{article} 
\usepackage{aaai2027}  
\usepackage[hyphens]{url}  
\usepackage{graphicx} 
\usepackage{natbib}  
\usepackage{caption} 
\usepackage{amsmath}
\usepackage{amssymb}

\usepackage{booktabs}
\usepackage{placeins}

\usepackage{tikz}
\usetikzlibrary{positioning,arrows.meta,fit,backgrounds,calc}

\title{Tracing the Unlabeled Storm: Cross-Variable Transfer in a Lagrangian
Atmospheric JEPA Framework}

\author {
    K M Anirudh\textsuperscript{\rm 1}\corresponding,
    S Sandeep\textsuperscript{\rm 1,\rm 3},
    Hariprasad Kodamana\textsuperscript{\rm 2,\rm 3}
}

\affiliations {
    \textsuperscript{\rm 1}Centre for Atmospheric Sciences, IIT Delhi, New Delhi, India\\
    \textsuperscript{\rm 2}Department of Chemical Engineering, IIT Delhi, New Delhi, India\\
    \textsuperscript{\rm 3}Yardi School of Artificial Intelligence, IIT Delhi, New Delhi, India\\
    anirudh@cas.iitd.ac.in
}

\begin{document}
\maketitle
\begin{abstract}
Deep atmospheric convection governs South Asian monsoon variability, yet attempting to learn its
latent world model directly from zero-inflated, heavy-tailed precipitation yields suboptimal
predictive representations. Continuous atmospheric proxies, such as outgoing longwave radiation
(OLR), express this convective organization far more coherently. We address this mismatch with
\emph{cross-variable proxy learning}: M-JEPA, a multiscale Monsoon Joint-Embedding Predictive
Architecture, is pretrained on five continuous proxy fields over Lagrangian patches tracking moving
convective systems---without rainfall supervision at any point. The resulting frozen representation
is transferred to daily precipitation forecasts through a shared decoder trunk featuring parallel
probabilistic and deterministic branches. Because rainfall is strictly unobserved during
pretraining, downstream skill directly measures the predictive information captured in the latent
rollout. A frozen-backbone probing framework with two controls (an identical architecture trained on
rainfall alone, and a randomly initialized backbone) attributes the transfer specifically to proxy
pretraining: direct rainfall training exhibits $36\%$ higher CRPS error ($7.52$ vs.\
$5.54$\,mm/day). Against the 51-member operational ECMWF ensemble, the transferred model attains a
statistically resolved CRPS advantage ($6.81$ vs.\ $6.89$\,mm/day) and higher Brier skill ($+0.05$
vs.\ $-0.04$) using $15.4$M parameters on a single consumer GPU, concentrated at heavy-rain
thresholds and fine spatial scales, while the ensemble retains an advantage in neighborhood skill
and deterministic references on point metrics. The result provides a competitive monsoon
precipitation forecast grounded in intraseasonal dynamics and a diagnostic framework for evaluating
transferred atmospheric representations.
\end{abstract}

\section{Introduction}
\label{sec:intro}
The South Asian monsoon (SAM) supplies the water on which nearly a quarter of the world's population depends, and its rainfall is among the most consequential and least predictable quantities in atmospheric science. Both physics-based numerical weather prediction (NWP) and modern data-driven forecasting systems continue to struggle with the processes governing its variability.

A fundamental challenge lies in precipitation itself. Unlike large-scale atmospheric variables, rainfall is sparse, intermittent, highly localized and strongly heavy-tailed, reflecting the nonlinear nature of deep convection. In NWP these processes occur below the grid scale and must be parameterized, a long-recognized source of systematic bias \citep{stephens2010}. Machine-learned weather models \citep{graphcast,pangu,aifs,aurora} achieve state-of-the-art skill on smooth continuous fields, but treat precipitation as one target among many, optimized with the same generic objectives despite its very different statistics.

The dominant source of subseasonal variability in the SAM is the Monsoon Intraseasonal Oscillation (MISO), a northward-propagating envelope of organized deep convection that governs the transition between active and break phases of the monsoon \citep{goswami2003,suhas2013}. Although rainfall constitutes the most visible manifestation of this evolution, the underlying convective organization is expressed far more coherently in continuous atmospheric fields, particularly outgoing longwave radiation (OLR). Rainfall superimposes fine-scale intermittency on this slowly evolving state, obscuring the organization that governs its occurrence. This motivates a cross-variable strategy in which the dynamics are learned from continuous proxies first and the representation is then transferred to rainfall.

Because the representation never sees rainfall, its ability to transfer to precipitation is a stringent test of the dynamics encoded in the latent. If a model pretrained only on continuous proxies predicts sparse, intermittent rainfall better than one trained directly on rainfall, the representation has captured transferable physical structure rather than task-specific shortcuts. Establishing that requires probing the representation itself, not downstream skill alone. Joint-Embedding Predictive Architectures (JEPAs) are widely motivated as learning latent ``world models''. What their rollouts encode has begun to be examined on video and robotics data, but not for atmospheric state, and not where the readout target is a physical variable absent from pretraining.

We address this question with \textbf{M-JEPA}, a multiscale autoregressive JEPA
trained in two stages. In the upstream stage, a context encoder and an exponential-moving-average
target encoder embed Lagrangian patches that follow tracked convective envelopes, and an
autoregressive predictor rolls the context latent forward seven daily steps under latent regression;
the five proxies are the only inputs, and rainfall is absent from the objective. In the downstream
stage, the encoder and predictor are frozen and the latent rollout is read out to daily precipitation
by a shared decoder trunk with a probabilistic and a deterministic branch, matched to rainfall's
zero-inflated statistics. Rainfall is therefore a transfer target throughout rather than a
pretraining objective, and every reported skill number is a measurement of the frozen representation.
Our contributions are fourfold:

\begin{itemize}
\item \textbf{A probing framework for atmospheric world models.} A frozen-backbone protocol
examining predictive information, variable-specific subspace organization, and temporal evolution
through Centered Kernel Alignment (CKA) (Section~\ref{sec:results}); a fourth probe,
latent-to-pixel attribution, is in supplementary Section~D.4.

\item \textbf{Evidence for transferable atmospheric dynamics.} Two controls, the identical
architecture pretrained on rainfall alone and a randomly initialized backbone, attribute the
transfer to learning on continuous proxies, not to architectural inductive bias or the decoder (Section~\ref{sec:controls}).

\item \textbf{Skill concentrated at spatially sharp extremes.} On matched windows M-JEPA Ens attains
a lower fair CRPS than the 51-member operational IFS Ens, by a resolved but small margin
($-0.08$\,mm/day, $95\%$ CI $[-0.16,-0.001]$), with higher Brier Skill Score, from a $15.4$M-parameter
model on one consumer GPU. The margin is not generic: member-FSS at the IMD Heavy and Very Heavy
thresholds grows with lead, and the fine-scale power spectrum tracks observations more closely than
IFS Ens's own members (Section~\ref{sec:readout}).

\item \textbf{Characterization of representation limitations.} The rollout is dynamically damped,
saturating by lead~5, and the decoded forecast under-continues the observed decay for the
longest-lived systems, motivating a scaling hypothesis in which faithful propagating dynamics need
proxy information denser in time, vertical levels and spatial extent (Section~\ref{sec:limits}).
\end{itemize}
\section{Related Work}

\paragraph{JEPA and latent world models.}
Joint-embedding predictive architectures predict in representation space rather than pixel space
\citep{ijepa,vjepa2}. M-JEPA follows V-JEPA~2 \citep{vjepa2} most directly, transposing its context
encoder, exponential-moving-average target and latent autoregressive rollout from video to a
co-moving atmospheric patch. Work asking what such rollouts encode
\citep{deltajepa,causaljepa,sdjepa} sits almost entirely in video and robotics, where the prediction
target shares the encoder's modality. We ask what a latent encodes when it is read out to a
\emph{different physical variable} than any it was pretrained on.

\paragraph{Machine-learned weather models.}
Emulators trained on reanalysis now match or exceed operational skill on the large-scale fields
\citep{graphcast,pangu,gencast,aifs}, pretrained models extend the recipe across variables and
resolutions \citep{aurora,climax}, and benchmarks have followed at medium and
subseasonal-to-seasonal range \citep{weatherbench2,chaosbench}. All are evaluated by forecast score
alone, leaving unexamined what their representations encode and whether it transfers off the
training variables. That is the question this paper puts to a JEPA latent, with GraphCast, AIFS and
the operational IFS as references.

\paragraph{Data-driven monsoon and precipitation prediction.}
Subseasonal prediction of the South Asian monsoon has largely proceeded by reducing the system
before predicting it: the oscillation is compressed to an index or a few oscillatory modes, these
are forecast, and rainfall is recovered from them \citep{anirudh2025miso,bach2024}. That keeps the
northward march of the envelope but compresses away the zonal structure determining where rain
falls; the representation here carries the evolving envelope field in longitude, latitude and time. Related efforts span Earth-system emulation of monsoon variability \citep{samudrace_eval},
subseasonal machine learning \citep{he2021subseasonal}, radar nowcasting
\citep{shi2015,wen2026duocast,lin2022clcrn}, dynamics-informed diffusion \citep{dyffusion},
quantile-aware subseasonal precipitation \citep{quantweather} and explicit heavy-tail distributions
\citep{naveau2016,wilson2022deepgpd}. All learn rainfall from rainfall; the representation studied
here never observes it.

\paragraph{Atmospheric latent representations.}
Atmospheric latents have been inspected directly \citep{betavae_pacific,latentda,weather2latent},
but always as \emph{reconstruction} latents of the same variables they encode. None studies a
self-supervised \emph{world-model} latent transferred to a held-out target. Frozen-encoder readouts
with probabilistic heads exist for precipitation \citep{dinov3crps}; that recipe is not claimed as
novel, and forecasting skill is treated here as a \emph{probe} of representation quality. The conjunction is new: a JEPA latent learned on object-centric \emph{Lagrangian patches},
transferred across variables to a held-out \emph{sparse} target, and analyzed with a
proxy$\rightarrow$latent$\rightarrow$pixel probing suite and controls attributing the transfer to the
proxies.

\begin{figure*}[t]
\centering
\includegraphics[width=0.62\textwidth]{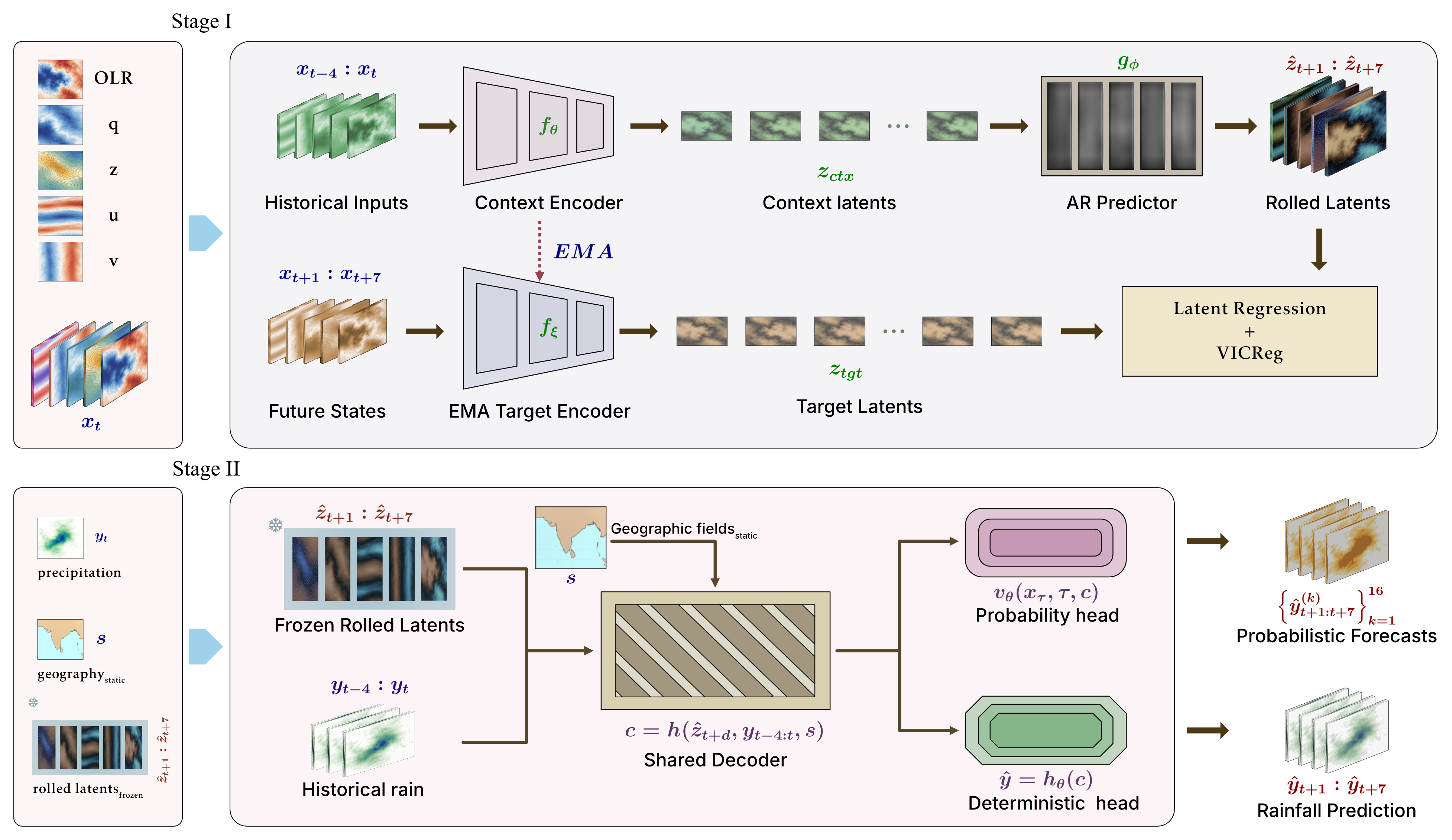}
\caption{The two stages of M-JEPA. \emph{Stage~I} (top): context encoder $f_\theta$, target encoder
$f_\xi$, autoregressive predictor $g_\phi$, pretrained on proxies with no rain objective.
\emph{Stage~II} (bottom): the frozen rollout $\hat z_{t+1:t+7}$, rain history $y_{t-4:t}$ and static
geography $s$ condition one shared decoder trunk whose two branches emit a $16$-member probabilistic
forecast and a deterministic field.}
\label{fig:arch}
\end{figure*}

\section{Data and Problem Setup}
\label{sec:data}

\paragraph{Approach.} We forecast the next seven daily precipitation fields over the South Asian
monsoon domain ($9.5^\circ$S--$29.5^\circ$N, $60.5$--$99.5^\circ$E) on a $1^\circ$ grid from five daily
proxy fields, with rainfall absent from pretraining. In place of a fixed Eulerian window the model
learns on \emph{Lagrangian patches} following one convective envelope at a time, so the encoder sees
a single system's evolution in its own co-moving frame.

\paragraph{Event tracking.} Envelopes are located by a classical detect-and-track pipeline on OLR
band-passed to the $20$--$90$ day intraseasonal band, which isolates the organized convective
envelope of the MISO from synoptic and seasonal variance \citep{wheeler1999,suhas2013}. Being retrospective and zero-phase, this filter selects and centers patches only; it is never a
model input, and the encoder consumes raw, unfiltered fields. Tracks are retained at a $12$-day minimum and tracking is direction-blind. Poleward motion is
therefore a property of the retained population ($60\%$ of test tracks), not a selection criterion. Because the
frame re-centers on the tracked centroid at every step, the envelope's bulk translation is supplied
by the tracker and is never predicted; the model forecasts intensity and structure within the moving
patch, and no metric credits it for knowing the system's future position. A sweep over the two tracker settings that fix patch geometry leaves the ordering against
persistence and deterministic IFS unchanged in all ten configurations, though not the ensemble
comparison, which we could not re-run per configuration (supplementary Section~E.5).

\paragraph{Data and windows.} The proxies are OLR \citep{lee2025olrcdr} and specific humidity $q$,
zonal and meridional wind $u,v$, and geopotential $z$ at 850\,hPa from ERA5
\citep{hersbach2020}; the verification target is GPM satellite rainfall \citep{huffman2015}. All
fields are regridded to the target grid, deseasonalized, and $z$-scored on training-year statistics
only. Each retained centroid anchors a tight foreground and a wider background crop, and every
five-day context plus seven-day horizon span along a track yields one window. Splits are by year
block: pretraining 1979--2017, decoder training 1998--2013 ($667$ windows, trained with
$\pm2$-grid-cell spatial-jitter augmentation), validation 2014--2017, test 2018--2025 ($337$
windows). Detection runs on all calendar months, so the pretraining population is intraseasonal
convective envelopes generally; evaluation is restricted to JJAS, when the northward-propagating
MISO envelope dominates that population. We impose none of the strict propagation filters used to
define MISO indices, so the retained tracks are a broader, on average slower set than a canonical
MISO composite (supplementary Section~A.1). Pretraining years overlap the decoder years, but both routes to leakage are closed: rainfall is
absent from every Stage-1 input and objective, and pretraining ends in 2017, before the first test year. Full details: supplementary Section~A.

\section{Method}
\label{sec:method}
M-JEPA is trained in two stages: self-supervised multiscale autoregressive JEPA pretraining on the
proxies (Stage~1), then a rainfall readout on the frozen latent rollout (Stage~2).
Figure~\ref{fig:arch} fixes the notation used throughout.

\subsection{Stage 1: upstream pretraining on proxies}
\label{sec:method-stage1}
Each input clip is a five-day sequence of Lagrangian patches $x_{t-4:t}$ anchored on the tracked
convective centroid, carried in two streams that differ in extent and token footprint, not in
grid spacing. The \emph{foreground} is OLR alone on a $20\times20$ patch at $1^\circ$, tokenized in
$4\times4$-cell tubelets; the \emph{background} carries $q,u,v,z$ at 850\,hPa plus a
spatially smoothed OLR envelope on a $40\times40$ patch about the same center, tokenized in
$8\times8$ tubelets. Both therefore yield a $5\times5$ grid of $256$-dimensional tokens, resolving
convective detail in the core and the synoptic-to-planetary environment around it at equal token
cost. Each stream is tokenized and tagged separately,
$u^{\mathrm{fg}}_d=\mathcal{E}^{\mathrm{fg}}(x^{\mathrm{fg}}_d)+e_{\mathrm{fg}}$ and
$u^{\mathrm{bg}}_d$ likewise, where the $\mathcal{E}$ are the per-stream tokenizers and
$e_{\mathrm{fg}},e_{\mathrm{bg}}$ are learned stream tags. The context encoder $T_\theta$, a
six-block eight-head pre-norm transformer, then embeds each day independently,
\begin{equation}
\label{eq:enc}
z_d = f_\theta(x_d) = T_\theta\big(\big[\,u^{\mathrm{fg}}_d \,\big\|\, u^{\mathrm{bg}}_d\,\big]\big)
\in\mathbb{R}^{50\times256},
\end{equation}
so the two scales interact through self-attention over the joint $50$-token set. All temporal
structure is left to the predictor. The context latents are the resulting five-day sequence
$z_{\mathrm{ctx}}=(z_{t-4},\dots,z_t)$.

The autoregressive predictor $g_\phi$ is a block-causal transformer (six blocks, eight heads) that
emits a residual increment and is applied to its own outputs to roll the state forward,
\begin{equation}
\label{eq:roll}
\begin{aligned}
\hat z_{t+d} &= \hat z_{t+d-1}+g_\phi\big(z_{\mathrm{ctx}},\,\hat z_{t+1:t+d-1}\big),\\
\hat z_t &\equiv z_t, \qquad d = 1,\dots,7,
\end{aligned}
\end{equation}
so each step conditions on the full sequence produced so far, and the zero-initialized output
projection starts training at persistence. Regression targets are supplied by the target encoder
$f_\xi$, an exponential-moving-average copy of $f_\theta$ that receives no gradient,
\begin{equation}
\label{eq:ema}
\begin{aligned}
z_{\mathrm{tgt},\,t+d} &= f_\xi(x_{t+d}),\\
\xi &\leftarrow \mu\,\xi+(1-\mu)\,\theta, \quad \mu = 0.996 .
\end{aligned}
\end{equation}
The Stage-1 objective combines latent regression in two forms with variance-covariance
regularization and an auxiliary reconstruction,
\begin{equation}
\label{eq:loss1}
\begin{aligned}
\mathcal{L}_{1}=\;&\textstyle\sum_{j}\,\ell\big(z_j+g_\phi(z_{\le j}),\,z_{\mathrm{tgt},\,j+1}\big)\\
&+\textstyle\sum_{d\le k}\,\ell\big(\hat z_{t+d},\,z_{\mathrm{tgt},\,t+d}\big)\\
&+\mathcal{V}(z)+0.04\,\mathcal{C}(z)\\
&+0.1\,\ell\big(D_{\mathrm{aux}}(z_d),\,x^{\mathrm{fg}}_d\big),
\end{aligned}
\end{equation}
where $\ell$ is the smooth-$L_1$ loss. The first term is teacher-forced and regresses every observed
one-day transition of the clip; the second is the free-running rollout at horizon $k$, raised on a
curriculum from two to seven days. $\mathcal{V}$ and $\mathcal{C}$ are the hinged unit-variance and
squared off-diagonal covariance terms of VICReg \citep{vicreg} that prevent latent collapse, and
$D_{\mathrm{aux}}$ is a reconstruction head that keeps the latent spatially decodable. Rainfall enters neither
\eqref{eq:loss1} nor any Stage-1 input. Trainable parameters are $4.8$M in $f_\theta$ and $4.9$M in
$g_\phi$; $f_\xi$ and $D_{\mathrm{aux}}$ are discarded after pretraining.

Both $f_\theta$ and $g_\phi$ are frozen for Stage~2, which makes the rollout a fixed operator rather
than a tunable feature extractor. Every probe and every forecast score therefore measures the
pretrained representation, not its adaptation to rainfall.

\subsection{Stage 2: downstream rainfall readout}
\label{sec:method-stage2}
Stage~2 reads three inputs per window: the frozen rollout $\hat z_{t+1:t+7}$ from \eqref{eq:roll},
the observed five-day rain history $y_{t-4:t}$ on the $20\times20$ target patch, and static
geography $s$ (land fraction, latitude, longitude). It has two parts, taken in turn below: a
shared trunk that decodes the rollout to output resolution, and two branches that read the trunk.
Writing $\hat z^{\mathrm{fg}}$ and
$\hat z^{\mathrm{bg}}$ for the foreground and background token blocks, the shared decoder trunk
projects both back onto the $5\times5$ token grid, fuses the two scales there with a $3\times3$
convolution, and upsamples to the output grid through two transposed-convolution blocks
($5\!\to\!10\!\to\!20$, $128$ channels),
\begin{equation}
\label{eq:trunk}
m_{t+d}=\mathcal{U}\big(W_{\mathrm{fuse}}\big[\,p^{\mathrm{fg}}_{t+d}
\,\big\|\,p^{\mathrm{bg}}_{t+d}\,\big]\big)
\in\mathbb{R}^{128\times20\times20},
\end{equation}
where $p^{\mathrm{fg}}_{t+d}=P_{\mathrm{fg}}(\hat z^{\mathrm{fg}}_{t+d})$ is the projected
foreground token block and $p^{\mathrm{bg}}_{t+d}$ likewise. Fusing at token resolution combines the
two scales while they are still spatially registered. Rain history and geography enter
afterwards at output resolution through their own convolutional encoders $E_y$, applied to
$\tilde y = \log(1+y_{t-4:t})$ with a learned token for windows that have no history, and $E_s$,
combined by $1\times1$ convolutions,
\begin{equation}
\label{eq:cond}
\begin{aligned}
c &= h(\hat z_{t+d},\,y_{t-4:t},\,s)\\
&= W_s\big[\,W_y\big[\,m_{t+d}\,\big\|\,E_y(\tilde y)\,\big]\;\big\|\;E_s(s)\,\big].
\end{aligned}
\end{equation}
The $E_y$ and $E_s$ maps are computed once per window and broadcast over leads, so all lead-to-lead
variation in $c$ originates in the rollout. A $1\times1$ readout $\hat o_{t+d}=W_{\mathrm{olr}}(m_{t+d})$
of the trunk feature also predicts the true future OLR field $o_{t+d}$, regularizing the trunk with
\begin{equation}
\label{eq:lolr}
\mathcal{L}_{\mathrm{OLR}}=\ell(\hat o_{t+d},\,o_{t+d}),
\end{equation}
the same smooth-$L_1$ $\ell$ as \eqref{eq:loss1}.

Two branches read $c$. The \emph{probability} branch is a
rectified flow with no deterministic anchor, unlike residual-correction designs such as CorrDiff
\citep{corrdiff}, and transports Gaussian noise to rain fields in square-root space: with
$x_1=\sqrt{y}$, $x_0\sim\mathcal{N}(0,I)$, $\tau\sim\mathcal{U}(0,1)$ and
$x_\tau=(1-\tau)x_0+\tau x_1$,
\begin{equation}
\label{eq:flow}
\begin{aligned}
\mathcal{L}_{\mathrm{fm}} &= \mathbb{E}\big\lVert v_\theta(x_\tau,\tau,c)-(x_1-x_0)\big\rVert^2,\\
\mathcal{L}_{\mathrm{es}} &= \mathrm{ES}\big(\{x^{(k)}\}_{k=1}^{4},\,x_1\big),
\end{aligned}
\end{equation}
the auxiliary term is the energy score \citep{gneiting2007} of four differentiable draws, scoring
whole sampled fields where $\mathcal{L}_{\mathrm{fm}}$ constrains only the velocity at one
interpolation time. The velocity field $v_\theta$ is a $20\!\to\!10\!\to\!5\!\to\!10\!\to\!20$ U-Net of width
$96$ ($3.0$M parameters) taking $x_\tau$ and a sinusoidal embedding of $\tau$, with $c$ pooled and
re-concatenated at every scale so the conditioning reaches the bottleneck undiluted. The trunk and
probability branch train jointly,
\begin{equation}
\label{eq:loss2}
\mathcal{L}_{2}=\mathcal{L}_{\mathrm{OLR}}+\mathcal{L}_{\mathrm{fm}}
+0.1\,\mathcal{L}_{\mathrm{es}} ,
\end{equation}
so the trunk's only training signal is this distributional objective.

The \emph{deterministic} branch $\hat y = h_\theta(c)$ shares the flow branch's U-Net topology at
width $32$ ($0.3$M parameters) but takes neither noise nor $\tau$, giving a single forward pass with a
softplus output. It is trained afterward, as a separate readout on the same trunk now held frozen,
on
\begin{equation}
\label{eq:det}
\mathcal{L}_{\mathrm{det}}=\mathcal{L}_{\mathrm{FACL}}(\hat y,y)+\lambda\,\mathcal{L}_{\mathrm{TDT}}(\hat y,y),
\qquad \lambda = 0.02,
\end{equation}
where, writing $\mathcal{F}$ for the 2-D orthonormal DFT over the spatial dims of a lead's field,
\begin{equation}
\label{eq:facl}
\mathcal{L}_{\mathrm{FAL}}=\mathbb{E}\big[(|\mathcal{F}\hat y|-|\mathcal{F}y|)^2\big], \quad
\mathcal{L}_{\mathrm{FCL}}=1-\frac{\mathrm{Re}\langle\mathcal{F}\hat y,\mathcal{F}y\rangle}
{\lVert\mathcal{F}\hat y\rVert\,\lVert\mathcal{F}y\rVert},
\end{equation}
and $\mathcal{L}_{\mathrm{FACL}}$ selects $\mathcal{L}_{\mathrm{FAL}}$ with probability
$p(\text{step})$, annealed $1\to0$ over the first $70\%$ of training, and
$\mathcal{L}_{\mathrm{FCL}}$ otherwise: spectral sharpness early, placement late. Writing $d\hat y_{t+d}=
\hat y_{t+d+1}-\hat y_{t+d}$ for the predicted lead-to-lead tendency and $dy_{t+d}$ likewise for
observed,
\begin{equation}
\label{eq:tdt}
\begin{aligned}
\mathcal{L}_{\mathrm{TDT}}&=\mathbb{E}\big[w_{t+d}\,\lvert d\hat y_{t+d}-dy_{t+d}\rvert\big],\\
w_{t+d}&=2 \text{ if } dy_{t+d}{<}0,\ d\hat y_{t+d}{>}dy_{t+d}, \text{ else } 1,
\end{aligned}
\end{equation}
i.e.\ an $L_1$ tendency penalty up-weighted by $2\times$ specifically when the forecast under-decays
relative to a truly decaying observation. Training the branches sequentially sidesteps the double-penalty problem structurally rather than by
loss weighting \citep{subich2025}, since a likelihood forecast and a squared-error point forecast are
different functionals of the predictive distribution, so one trunk optimized for both at once is
optimal for neither. Trunks trained under $\mathcal{L}_{\mathrm{det}}$ itself top out at pooled ACC
$0.255$--$0.271$ regardless of curriculum, whereas reading out from the frozen
distributionally-trained trunk reaches $0.292$ on that same pooled convention; the corresponding
case-averaged scores, which are the convention used for every result we report, are in
Section~\ref{sec:sharpness}. Across Stage~2 the trunk holds $2.1$M parameters and the conditioning encoders $0.4$M, for $5.7$M
against the $9.7$M frozen Stage-1 stack (components rounded independently, summing to $5.8$M).

Each window therefore yields, at every lead $d=1,\dots,7$, a $16$-member probabilistic rainfall
field $\{\hat y^{(k)}_{t+1:t+7}\}_{k=1}^{16}$, obtained by integrating $v_\theta$ from independent
noise and squaring back to mm\,day$^{-1}$, and one deterministic field $\hat y_{t+1:t+7}$, both used
directly with no post-hoc recalibration. Architecture and optimization
settings for both stages: supplementary Section~B.

Every score below is computed in the patch frame and case-averaged, using the fair unbiased CRPS
estimator \citep{ferro2008} and the NHC homogeneous-sample matching rule; these conventions, and
member-FSS, Brier skill and ACC, are defined once in supplementary Section~C and then used without
restatement.

\section{Results}
\label{sec:results}

We report forecast skill against the operational references first, then what the frozen
representation contains that makes that skill possible. Token- and pixel-level attribution is
deferred to supplementary Section~D.4.

\subsection{Probabilistic forecast skill}
\label{sec:readout}
The probabilistic reference is IFS Ens, the $51$-member operational ECMWF ensemble, and the same
underlying model as the deterministic IFS run of Section~\ref{sec:sharpness}. On matched windows
($n{=}169$), M-JEPA Ens attains CRPS $6.81$ against $6.89$\,mm/day and positive Brier skill ($+0.05$
against $-0.04$), from a $15.4$M-parameter model on one $16$\,GB consumer GPU. The CRPS margin is
resolved but slight: $\Delta=-0.08$, $95\%$ CI $[-0.16,-0.001]$ over $n{=}1183$ paired cases. In
exchange, IFS Ens keeps a clear advantage on neighborhood skill (Table~\ref{tab:readout}), the first
appearance of a sharpness-versus-overlap trade-off that recurs throughout.

\begin{table}[t]
\centering\small
\caption{Forecast skill on matched windows against the $51$-member operational IFS ensemble
($n{=}169$, fair CRPS in mm/day; the IFS (det.) row is on its own single-member convention). The CRPS difference against IFS Ens is $-0.08$\,mm/day, $95\%$ paired-bootstrap CI
$[-0.16,-0.001]$ over $10^4$ resamples. Boldface marks the better ensemble. Methodology:
supplementary Section~C.}
\label{tab:readout}
\resizebox{\columnwidth}{!}{%
\begin{tabular}{lcccccc}
\toprule
 & Members & CRPS $\downarrow$ & BSS $\uparrow$ & FSS10 $\uparrow$ & Spread/RMSE $\to1$ & Rank dev.\ $\downarrow$ \\
\midrule
M-JEPA Ens & 16 & \textbf{6.81} & \textbf{+0.05} & 0.53 & 0.73 & \textbf{0.60} \\
IFS Ens & 51 & 6.89 & $-0.04$ & \textbf{0.63} & 0.45 & 0.95 \\
IFS (det.) & 1 & 9.76 & $-0.36$ & 0.73 & 0.00 & 0.70 \\
\bottomrule
\end{tabular}}
\end{table}

\begin{figure}[t]
\centering
\includegraphics[width=\linewidth]{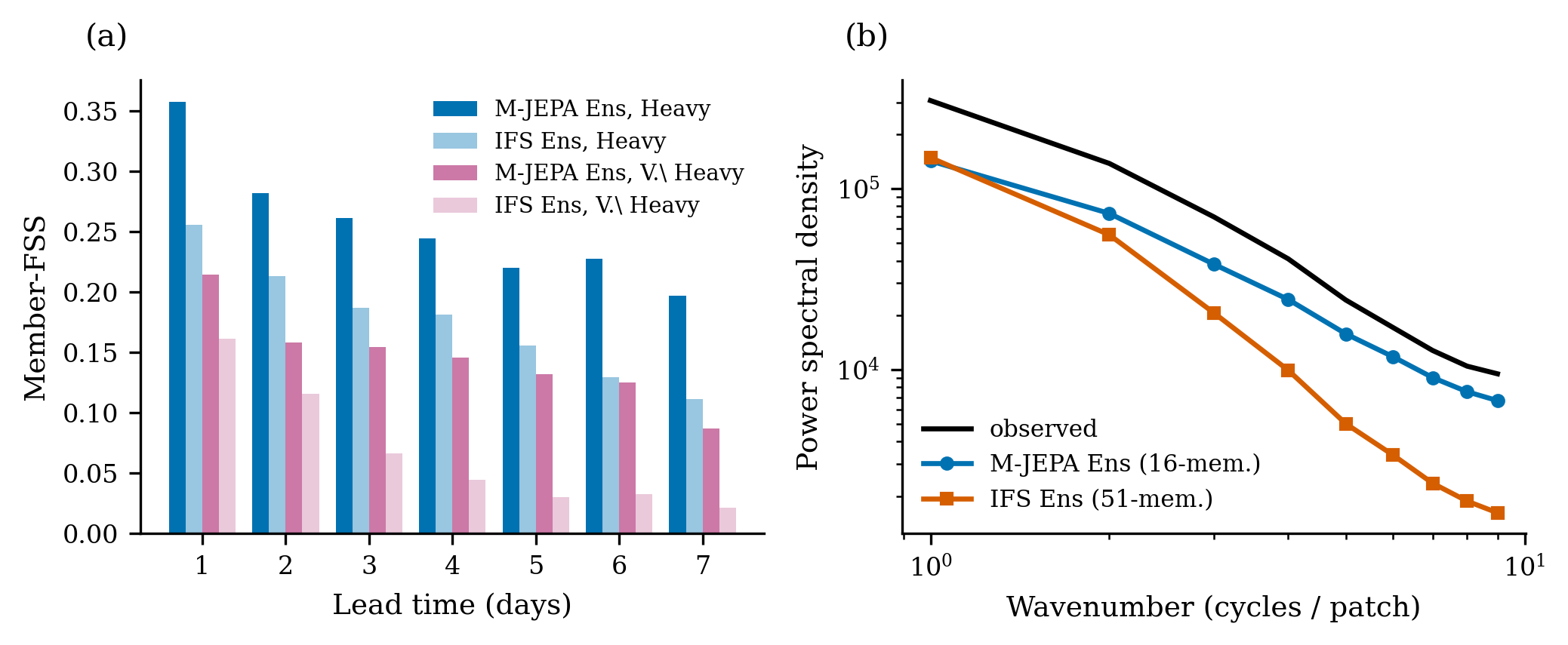}
\caption{Probability-branch diagnostics, matched $n{=}169$ population.
\emph{(a)} Member-FSS at IMD Heavy/Very Heavy thresholds by lead. \emph{(b)} Radial power spectral
density, M-JEPA Ens vs.\ IFS Ens vs.\ observed.}
\label{fig:prob-diag}
\end{figure}

Figure~\ref{fig:prob-diag}a disaggregates M-JEPA Ens's member-FSS advantage over IFS Ens (matched
$n{=}169$) by lead, at the India Meteorological Department's operational 24-hour rainfall
classification \citep{falga2022}, Heavy ($64.5$\,mm) and Very Heavy ($115.6$\,mm); the margin
is larger at long leads than at short (Heavy: $1.4\times$ at lead~1 to $1.8\times$ at lead~7;
Very Heavy: $1.3\times$ to $4.1\times$), a placement-tolerant, neighborhood-skill effect distinct
from raw point error at extreme intensities. At the third IMD tier, Extremely Heavy
($\geq204.5$\,mm, not shown), the ordering is lead-dependent (M-JEPA first in each pair): IFS Ens leads at the first two leads
($0.069$ vs.\ $0.196$; $0.062$ vs.\ $0.144$), M-JEPA Ens from lead~3 onward by up to $6.6\times$ at
lead~5 ($0.046$ vs.\ $0.007$), on $27$--$38$ events per lead; pooled, the short-lead deficit
dominates the tier ($0.056$ vs.\ $0.077$).

Figure~\ref{fig:prob-diag}b gives a candidate mechanism: the radial power spectral density of each
ensemble's individual members against observations, pooled over the matched population and all seven
leads, reported descriptively rather than as a hypothesis test. Both systems miss the observed spectrum, but in opposite
directions as wavenumber increases. IFS Ens is closest at the largest scale ($48\%$ of observed
power) and falls to $17\%$ at the finest resolved wavenumber, the smoothing signature in which
mass-conserving blur spreads a sharp feature over a wider area at lower amplitude. M-JEPA Ens starts no
better ($46\%$) but its share \emph{rises} with wavenumber to $71$--$72\%$ at the finest scales, four
times IFS Ens's retention there. Matching fine-scale power shows only that the right \emph{amount} of small-scale variance survives,
not that it is placed correctly. Panel~(a)'s member-FSS margin, which does credit placement, is the
complementary evidence. Estimator and per-radius values:
supplementary Section~F.6.

\subsection{Deterministic skill and metric sensitivity}
\label{sec:sharpness}
Case-averaged over the test set ($n{=}337$ windows, seven leads), the deterministic branch attains
RMSE $16.24$\,mm/day and spatial anomaly correlation $0.31$ against persistence's $23.46$ and $0.18$,
exceeding it on both at every lead; persistence nonetheless retains a small advantage at the
$3^\circ$ and $5^\circ$ neighborhood widths ($0.49/0.56$ against $0.47/0.53$), the same
sharpness-versus-overlap trade-off that separates M-JEPA Ens from IFS Ens. Against the operational archives, restricted to the
cells the tracked patches cover and each on its own matched-coverage subset ($266/127/127$ distinct
target days, from $161/66/66$ tracked windows),
M-JEPA's RMSE is lower than IFS's ($18.23$ vs.\ $18.45$, leading at six of the seven leads); on ACC
and FSS against all three references, and on RMSE against AIFS and GraphCast, the references lead
(supplementary Table~S12).

This split reflects metric sensitivity as much as skill: under a scale-invariant ACC a forecast that
sheds fine-scale detail can score well by being smooth, and both learned archives show the signature. AIFS self-smooths progressively over the horizon (roughness ratio $0.396\to0.323$); GraphCast is
uniformly over-smooth from lead~1 and gains instead as the truth simplifies, observed roughness
falling $22\%$ as the tracked system ages. At case level each (window, lead) case's ACC correlates
negatively with its own observed roughness (Pearson $r=-0.28$ IFS, $-0.57$ AIFS, $-0.57$ GraphCast),
so rougher targets score lower regardless of lead (supplementary Section~F.2). The FSS margin carries a further
confound, since the archives are regridded from $0.25^\circ$ onto our $1^\circ$ grid, smoothing them
before scoring. Two structural asymmetries remain regardless. Each reference initializes from an analysis-quality
state at 6-hourly or finer cadence \citep{graphcast,aifs}, against a frozen daily latent; and each
forecasts the full domain, against one co-moving patch.

\begin{figure}[t]
\centering
\includegraphics[width=\linewidth]{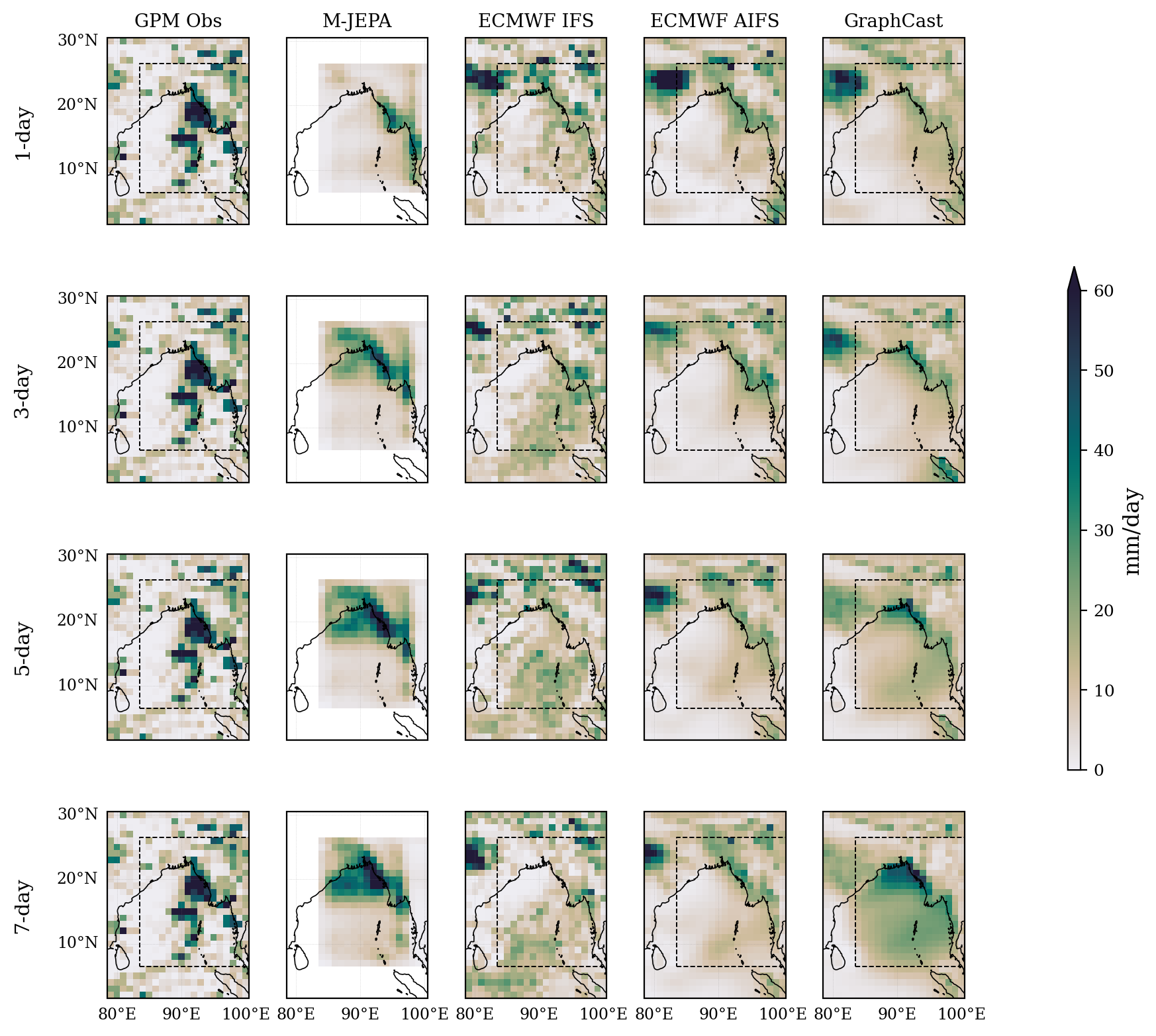}
\caption{Five-model spatial comparison on a matched target day (2024-08-04). Columns: GPM
observations, M-JEPA, IFS, AIFS, GraphCast; rows: forecasts issued $1$, $3$, $5$, $7$ days
ahead. Operational fields are single deterministic members cropped at the same Lagrangian patch, so
this compares placement, not probabilistic skill; dashed box marks the native patch extent.}
\label{fig:spatial5model}
\end{figure}

On metrics that score sharpness directly the ordering reverses. Measured by dry-area specificity,
the correct-negative rate on truly-dry cells at the $2$\,mm threshold, the deterministic branch
exceeds IFS and GraphCast at every lead and AIFS at six of seven, losing only at lead~1
(Table~\ref{tab:specificity}). Figure~\ref{fig:spatial5model} shows the same contrast on a
single day.

\begin{table}[t]
\centering\small
\caption{Dry-area specificity (correct-negative rate on truly-dry cells, $2$\,mm threshold) by
lead. AIFS and GraphCast share one matched subset, so a single M-JEPA row serves both.}
\label{tab:specificity}
\resizebox{\columnwidth}{!}{%
\begin{tabular}{lccccccc}
\toprule
System $\uparrow$ & 1 & 2 & 3 & 4 & 5 & 6 & 7 \\
\midrule
M-JEPA (IFS subset) & \textbf{0.555} & \textbf{0.534} & \textbf{0.516} & \textbf{0.503} & \textbf{0.503} & \textbf{0.492} & \textbf{0.475} \\
IFS & 0.540 & 0.477 & 0.460 & 0.452 & 0.462 & 0.463 & 0.454 \\
\midrule
M-JEPA (AIFS/GC subset) & 0.516 & \textbf{0.532} & \textbf{0.521} & \textbf{0.493} & \textbf{0.487} & \textbf{0.480} & \textbf{0.457} \\
AIFS & \textbf{0.526} & 0.497 & 0.479 & 0.471 & 0.443 & 0.427 & 0.430 \\
GraphCast & 0.424 & 0.414 & 0.400 & 0.387 & 0.378 & 0.366 & 0.377 \\
\bottomrule
\end{tabular}}
\end{table}

\subsection{Event lifecycle and amplitude evolution}
\label{sec:probe-evolution}
The decoded forecast tracks a convective event's evolution, not merely its instantaneous intensity,
on two independent axes. On \emph{position}, comparing the \emph{sign} of the precip-weighted
centroid's drift relative to lead~1, the forecast matches the observed direction more often than the
$50\%$ chance rate at every lead (pooled $60.5\%$ northward and $59.0\%$ zonal, $n{=}2022$, binomial
$p<10^{-4}$ on both axes), rising to $65.0\%/64.0\%$ once near-zero observed movement is excluded, so
the signal strengthens rather than weakens on genuine motion. M-JEPA still clears chance on the $n{=}161$ subset where IFS also has a matched forecast ($60.6\%$
N/S, $56.1\%$ E/W, $p<10^{-3}$). IFS is nonetheless the more accurate of the two there ($68.0\%$,
$59.0\%$), as its full-physics initialization and sub-daily cadence would predict. On \emph{amplitude}, the model reproduces the decay both across the forecast horizon
(lead~7 relative to lead~1: $0.74\times$ forecast against $0.73\times$ observed) and across an
event's own lifecycle ($-0.0253$ per day against $-0.0247$ observed, a $2\%$ over-decay, on a dry
bias of $2.1$\,mm/day). Reproducing lifecycle decay matters here because it is the same aging signal
that inflates the archives' ACC: the model follows the observed simplification rather than being
scored by it.

Against those two successes the latent's own trajectory is more limited. Linear Centered Kernel Alignment
\citep[CKA;][]{kornblith2019} between the context and rolled latents falls from $0.94$ at lead~1 to
$0.79$ by lead~5 and then plateaus, so the rollout transports the representation rather than copying
it forward while its trajectory saturates (supplementary Section~D.3). Consistently, beyond day~16,
where the sample thins to the longest-lived tracks, the observed composite continues falling to
$0.68\times$ its day-5 value while the decoded composite stops at $0.75\times$, so the forecast
under-continues the decay of the longest-lived systems (supplementary Section~F.5).

\subsection{Latent variable structure}
\label{sec:probe-variables}
The five proxies survive pretraining in a readable, physically organized form. Linear probes on the
token-averaged context latent recover each proxy's patch-mean with $R^2$ between $0.67$ (zonal wind)
and $0.85$ (geopotential), and the content is \emph{linearly} accessible rather than hidden in a
nonlinear code, since a multilayer probe tuned by cross-validation, with model selection never
touching the test years, improves on the linear probe by at most $+0.02$, on none of the five significantly
(supplementary Section~D.2). The five probe directions are also close to mutually orthogonal (mean
absolute off-diagonal cosine $0.15$), and the two largest overlaps are physically interpretable:
geopotential with the OLR envelope ($+0.38$) and humidity with the meridional wind ($+0.30$),
corresponding to convection organizing beneath lowered geopotential and to the low-level southerly
flow that moistens the envelope's poleward flank, recovered without variable labels.

The same latent separates extreme-rain windows before any decoding. A logistic probe detecting
whether a window reaches a widespread-heavy-rain day (area-mean $\geq 15$\,mm/day at any lead, $27\%$
base rate, $n{=}337$) attains ROC-AUC $0.67$ and average precision $0.41$. Modest but real, that separation locates the extreme-skill result of Section~\ref{sec:readout}
partly in the representation, not solely in the decoder.

\subsection{Predictive information in the latent rollout}
\label{sec:probe-predinfo}
To test whether the rollout is a genuine \emph{forecast} rather than a re-encoding of the present, we
regress area-mean rainfall on day $t{+}d$ from the rolled latent $\hat z_{t+d}$ and from a
same-capacity control, the frozen context-end latent $z_t$ reused at every lead, which would match it
if the rollout added nothing. Instead the rolled latent decodes future rainfall better at \emph{every} lead (pooled $R^2$ $0.174$ against $0.084$, margin $+0.090$, $95\%$
paired-bootstrap CI $[+0.011,+0.178]$, $10^4$ resamples; per-lead breakdown, supplementary
Section~D.1), peaking at lead~3 ($+0.139$, where the static probe falls below the unconditional-mean
baseline) and decaying to $+0.048$ by lead~7. Individually, only the pooled margin and lead~1 exclude zero: the per-lead intervals are wide at
this test-window count, so the claim rests on the pooled statistic. Even so, the rollout carries decodable information about the
future that the present-state encoding lacks, and is therefore predicting rather than
autoencoding.

\begin{table}[t]
\centering\small
\caption{Isolating the source of transfer. The first two rows share architecture, recipe and
probability head, so only the pretraining \emph{signal} differs; the last three are conventional
networks given the same rainfall history. CRPS in mm/day, no recalibration, on the
full test set ($n{=}337$) except the rain-only control, which rainfall sparsity leaves a smaller
held-out population ($n{=}118$). Table~\ref{tab:readout}'s $6.81$ is this same model on the
$n{=}169$ windows the IFS ensemble also covers.}
\label{tab:controls}
\begin{tabular}{lrcc}
\toprule
Pretraining signal & Params & CRPS $\downarrow$ & BSS $\uparrow$ \\
\midrule
Five proxies & 9.7M & \textbf{5.54} & \textbf{+0.18} \\
Rainfall only, same recipe & 9.6M & 7.52 & $-0.04$ \\
\midrule
U-Net (rain history) & 351K & 7.38 & $-0.62$ \\
ConvLSTM (rain history) & 22K & 8.33 & $-1.07$ \\
ViT (rain history) & 836K & 8.65 & $-0.99$ \\
\bottomrule
\end{tabular}
\end{table}

\subsection{Attribution of the transfer}
\label{sec:controls}
Four checks attribute the foregoing to the pretraining \emph{signal} rather than the architecture
(supplementary Section~E): a randomly initialized backbone of identical architecture (trained
exceeds random by $+0.128$ pooled $R^2$), a second held-out target in future OLR (margin $+0.126$),
a leave-one-proxy-out ablation in which meridional wind is the only individually load-bearing
proxy, and a tracker-sensitivity sweep.

The principal control replaces the pretraining signal itself. Applying the identical JEPA architecture
and recipe to the rainfall field, holding everything else fixed, raises CRPS error by $36\%$ ($7.52$ against
$5.54$\,mm/day) and collapses Brier skill from $+0.18$ to $-0.04$ (Table~\ref{tab:controls}); the two
are indistinguishable only on neighborhood skill (FSS10 $0.53$ and $0.54$). The mechanism is data
sparsity: rainfall is intermittent, so the same tracking yields only $226$ valid pretraining windows
against $3885$ for the all-season proxies. Even so, the control outperforms three conventional
rain-history networks on the same patches and splits, all of which have \emph{negative} Brier skill
across two orders of magnitude in parameter count. What makes the representation transfer is
therefore the proxy choice, not the architecture or the parameter budget.

\section{Limitations and Scaling Outlook}
\label{sec:limits}

Our primary limitation concerns temporal rollout dynamics rather than representation capacity. The
latent trajectory saturates by lead~5 (Section~\ref{sec:probe-evolution}), so decoded forecasts
under-predict the observed decay of long-lived systems. Such damping is expected under squared-error
regression, which contracts toward the conditional mean as uncertainty grows; isolating
loss-induced damping from representational limits would require an ensemble-valued rollout.

The Lagrangian frame also bounds the benchmark scope. Metrics are evaluated only over tracked
convective patches, with operational references cropped into co-moving frames, and admitting only
tracked systems caps the test set at $337$ windows and $169$ matched days on a single seed. The
headline CRPS advantage is resolved on this sample; broader generalization requires further evaluation.

These constraints indicate an information-limited rather than an idea-limited regime. Under a
$17\times$ reduction in pretraining data volume the rain-only control degrades gracefully, still
outperforming three conventional rain-history networks (Section~\ref{sec:controls}). Overcoming
lead-5 saturation therefore calls for denser proxy inputs, at higher temporal frequency and over
more vertical levels and a broader domain, rather than a larger parameter count (supplementary
Section~G).

\section{Conclusion}
\label{sec:conclusion}

Self-supervised pretraining on continuous atmospheric proxies enables M-JEPA to learn transferable
subseasonal dynamics without rainfall supervision. Across a seven-day horizon, the frozen latent
rollout decodes future rainfall better than a static present-state baseline (pooled $R^2$ margin
$+0.090$), whereas direct precipitation training incurs $36\%$ higher CRPS error ($7.52$ vs.\
$5.54$\,mm/day), confirming the value of proxy representations.

Against the $51$-member ECMWF operational ensemble, M-JEPA attains a small but resolved advantage in
fair CRPS ($6.81$ vs.\ $6.89$\,mm/day) and positive Brier skill ($+0.05$ vs.\ $-0.04$). These gains
concentrate at heavy rainfall thresholds (member-FSS advantage growing $1.4\times \to 1.8\times$ over
the horizon) and fine spatial scales ($71$--$72\%$ spectral power retention vs.\ $17\%$), while the
ensemble retains the advantage on neighborhood skill and the deterministic references on point
metrics. Taken together, this demonstrates that latent cross-variable
learning can capture physical dynamics across disjoint targets.

\bibliography{references}

\clearpage

\renewcommand{\thetable}{S\arabic{table}}
\renewcommand{\thefigure}{S\arabic{figure}}
\renewcommand{\theequation}{S\arabic{equation}}
\setcounter{table}{0}\setcounter{figure}{0}\setcounter{equation}{0}

\begin{center}
{\Large\bfseries Supplementary Material}
\end{center}
\renewcommand{\thesection}{\Alph{section}}
\setcounter{section}{0}


\section{Data and Problem Setup}
\label{sec:supp-data}

We predict the next seven daily precipitation fields over the South Asian monsoon domain
($9.5^\circ$S--$29.5^\circ$N, $60.5$--$99.5^\circ$E) on a $1^\circ$ grid. Inputs are five daily proxy
fields: outgoing longwave radiation (OLR) plus specific humidity, zonal and meridional wind, and
geopotential at 850\,hPa \citep{hersbach2020}. The target is satellite precipitation
\citep{huffman2015}. All fields are regridded to the target grid, deseasonalized, and $z$-scored
using training-year statistics only. Rather than a fixed Eulerian window, we extract
\emph{Lagrangian patches} that follow convective systems identified by a classical
detection-and-tracking pipeline on band-passed OLR; the full detection and tracking algorithm and
its settings are given in Section~\ref{sec:supp-tracking}, and the trajectories of the resulting
test population in Figure~\ref{fig:supp-all-tracks}. This retrospective, zero-phase
(non-causal) bandpass filter is used \emph{only} to define patch geometry (which convective
system to follow, and where to center a window), sharing only the 20--90 day frequency band with
the convention used to define standard real-time MISO monitoring indices \citep{suhas2013}, not
their causal (forward-only) design; the tracking filter itself is never used to construct a model
input value. The model forecasts the \emph{intensity and structural evolution} of
rainfall within the patch, not the patch trajectory. The tracker-supplied center coordinate fixes
where each evaluation window sits but is never a predicted quantity, so it acts as an isolated oracle
for window placement, and no forecast metric credits the model for knowing the system's future
position. All five daily proxy fields the encoder consumes are raw and unfiltered; an
optional bandpass-filtered auxiliary channel exists in a strictly causal (forward-only) variant but
is disabled by default. A tight foreground patch captures the convective core; a wider background
patch captures the environmental circulation, with a purely spatial (not temporal) Gaussian
smoothing of OLR that carries no information from other time steps. The analysis domain is itself
$40\times40$ cells, the same nominal size as the background patch, so the co-moving crop is taken
from a zero-padded field extended by half a patch width on every side rather than from the raw
domain. A background patch therefore remains centered on the tracked centroid wherever the system
sits, and coincides with the domain only for a system at dead center; for systems near an edge the
out-of-domain portion is zero, which after deseasonalizing and $z$-scoring is the climatological
mean rather than an artificial extremum. The foreground patch, at $20\times20$, is well inside the
domain for all retained tracks. The sensitivity of the
comparative conclusions to the two tracker-geometry parameters is established separately in
Section~\ref{sec:supp-tracker}.

\subsection{Object Detection and Tracking}
\label{sec:supp-tracking}

The detect-and-track pipeline runs on daily OLR over the all-variable box domain, all calendar
months, at $1^\circ$ resolution.

\paragraph{Signal.} Each pixel's OLR time series is band-pass filtered to the $20$--$90$ day
intraseasonal band with a third-order Butterworth filter applied zero-phase (forward--backward),
the standard band for isolating the intraseasonal envelope \citep{wheeler1999}; missing values are
linearly interpolated before filtering. This retrospective filter defines patch geometry only and
never enters a model input.

\paragraph{Detection.} On each day, deep-convective cores are the strongly negative anomalies:
cells below the $10$th-percentile quantile of that day's band-passed field are selected,
Gaussian-smoothed ($\sigma{=}1$ cell) to bridge fragmented cores, and grouped into connected
components (\texttt{scipy.ndimage.label}); components smaller than $30$ cells are discarded. Each
object's location is its \emph{intensity-weighted} centroid, each cell weighted by its absolute OLR
anomaly, so the center follows the deepest convection and not the surrounding stratiform anvil.

\paragraph{Tracking.} Detections are linked into tracks by greedy nearest-cost assignment within
each year. Matching object $j$ on day $t$ to an existing track $i$ costs
\begin{equation*}
c_{ij} = w_d\,\lVert \mathbf{x}_j - \hat{\mathbf{x}}_i\rVert
       + w_a\,\frac{|A_j-A_i|}{A_i}
       + w_t\,(\Delta t - 1)
       - w_n\,\Delta\mathrm{lat},
\end{equation*}
with distance weight $w_d{=}1.0$, fractional-area-change weight $w_a{=}0.5$, time-gap weight
$w_t{=}1.0$, and direction weight $w_n{=}0$. The direction term is identically zero, because tracking is
direction-blind, so northward propagation is never rewarded in the assignment cost. Here $\hat{\mathbf{x}}_i$ is the
track position linearly extrapolated from its last step, which bridges up to four missed detections
($\mathrm{gap\_days}{=}4$); candidate matches are restricted to a $5^\circ$/day $\times\,\Delta t$
search radius.

\begin{figure*}[t]
\centering
\includegraphics[width=0.95\textwidth]{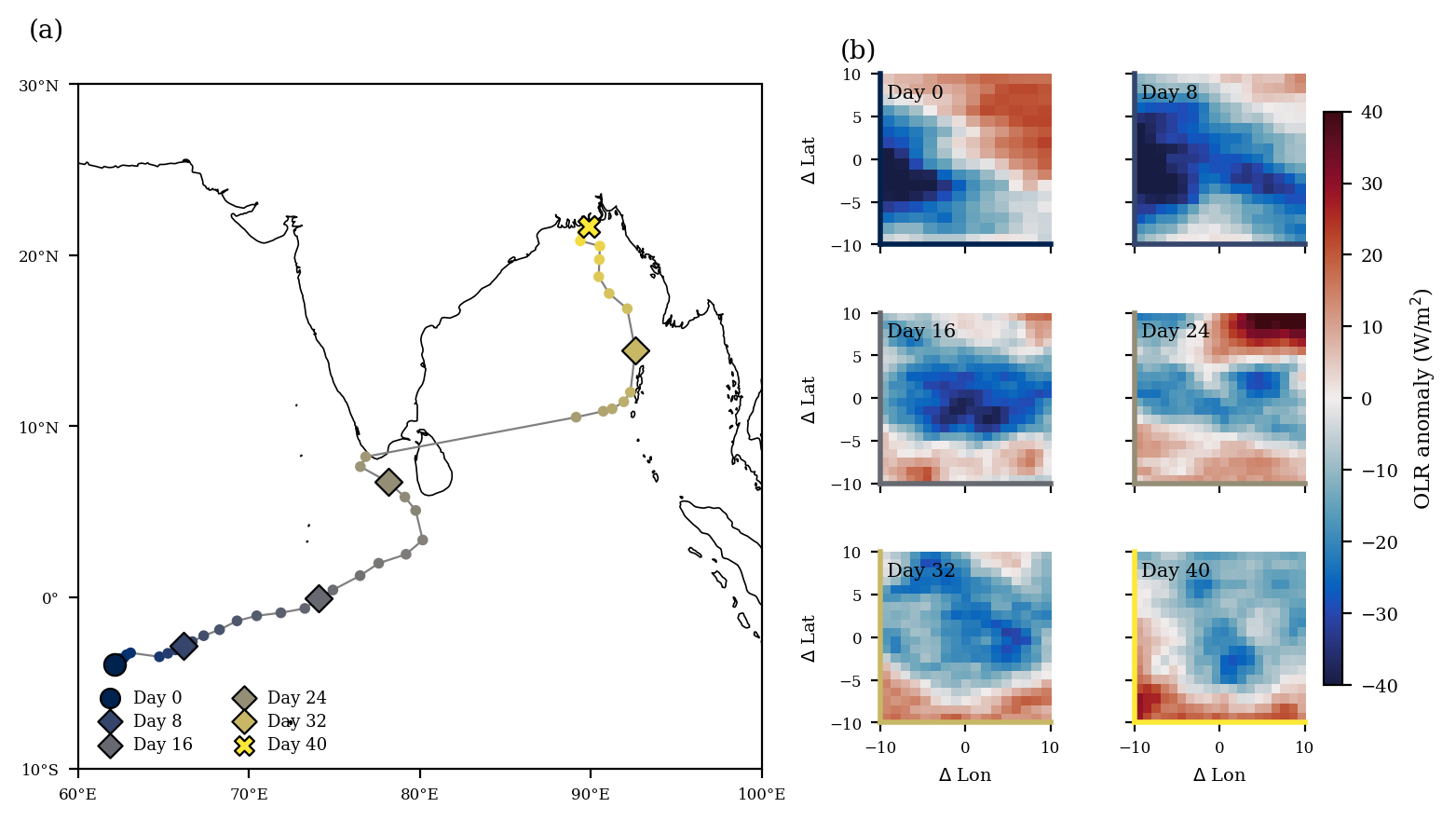}
\caption{Lagrangian tracking of one long-lived convective envelope (track 2766, genesis
2021-09-07, ocean, equatorial Indian Ocean; 45-day duration; net northward displacement
$+25.6^\circ$). (\textbf{a}) The tracked centroid trajectory from genesis to terminus, colored by
elapsed time; diamonds mark the six days shown at right. (\textbf{b}) The $20^\circ\times20^\circ$
bandpassed OLR anomaly patch centered on the moving centroid at each marked day, its border colored
to match that day on the trajectory, illustrating how
the Lagrangian frame re-centers on the system as it propagates from the equatorial Indian Ocean to
the Bay of Bengal. This track is illustrative rather than typical, since the retained test population's
median net northward displacement is $+3.5^\circ$. This patch geometry, not the raw OLR values shown here, is what the tracker
supplies; the encoder itself consumes unfiltered fields (Section~A).}
\label{fig:supp-track-example}
\end{figure*}

\begin{figure}[t]
\centering
\includegraphics[width=0.9\linewidth]{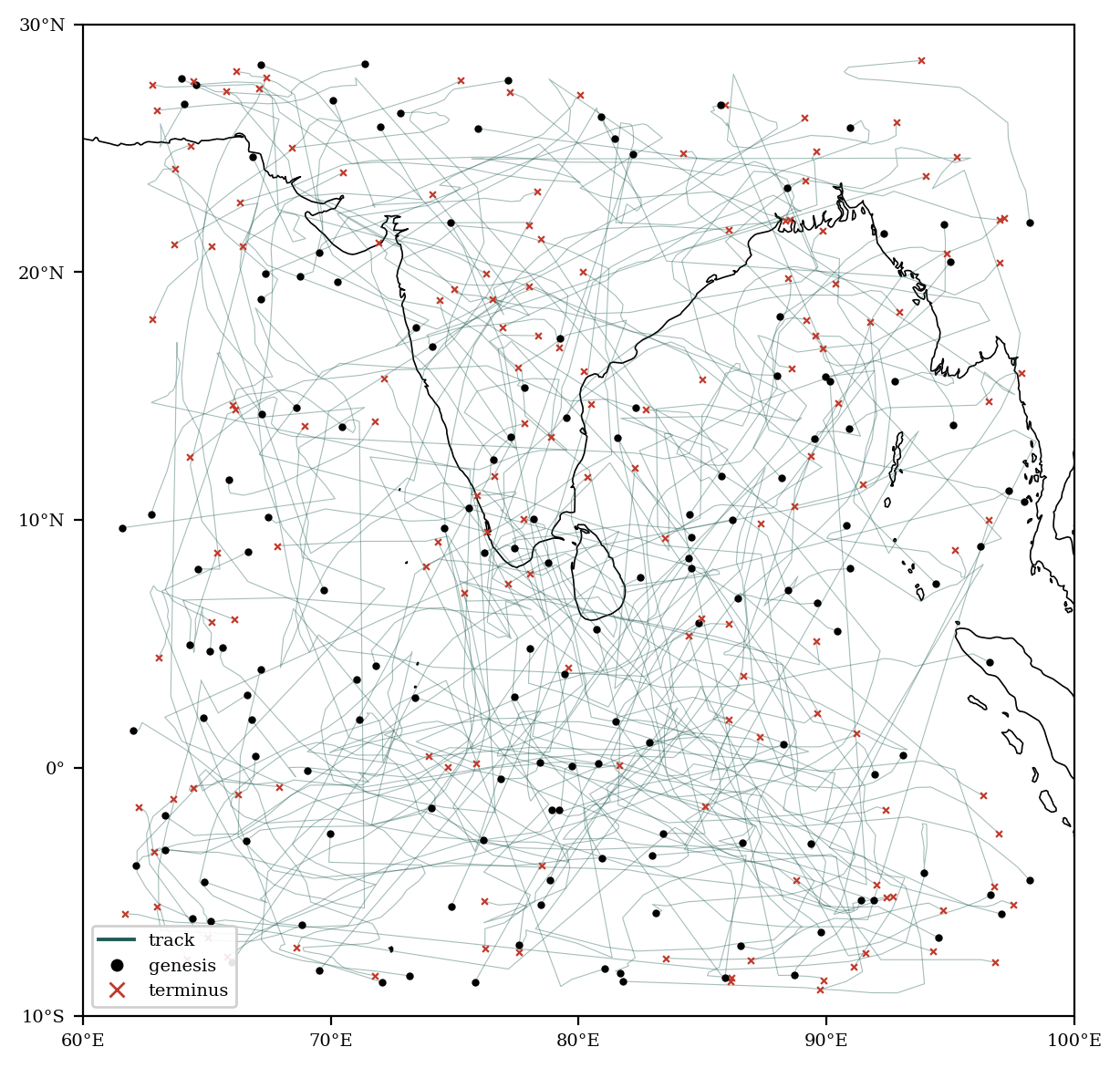}
\caption{Lagrangian trajectories of all 149 convective OLR objects tracked in the held-out test
population (genesis years 2018--2025, JJAS), the population underlying every forecast-skill number
reported in this paper. Black circles mark genesis and red crosses mark terminus, with genesis
concentrated across the equatorial Indian Ocean and Bay of Bengal; net northward displacement is
positive on balance (median $+3.5^\circ$, Section~A).
The 1979--2017 self-supervised pretraining population (711 tracks, of which the 2014--2017 subset
also serves as validation) shows the same qualitative spatial pattern.}
\label{fig:supp-all-tracks}
\end{figure}

\paragraph{Retention.} A track is kept if it spans at least $12$ days, the five-day context plus
seven-day forecast horizon. The strict-MISO physical filters of the original tracker (minimum net
northward displacement, minimum and maximum propagation speed, and an ocean-genesis mask) are
disabled here, which retains a larger and more varied training population. The test-year
population ($149$ tracks, before the span filter) has median net northward displacement
$+3.5^\circ$, or under
$0.3^\circ$ per day over the retained window, with $60\%$ of tracks propagating poleward ($2.4$
binomial standard errors above chance at this sample size). Poleward motion therefore emerges from
the data and not from the direction-blind tracking cost, as a tendency and not a selection
criterion, and the median track is slower than a canonical MISO envelope. Each retained centroid
then anchors the two-scale (tight foreground, wide background) co-moving crop described above.
Figure~\ref{fig:supp-track-example} follows one long-lived envelope through this procedure, and
Figure~\ref{fig:supp-all-tracks} shows the trajectories of the full test-year population.

Splits are by year block to avoid autocorrelation leakage: self-supervised pretraining on
1979--2017, decoder training on 1998--2013 (667 windows, trained with $\pm2$-grid-cell
spatial-jitter augmentation), validation on 2014--2017, and a held-out test set of 2018--2025 (337 windows).

\section{Model Architecture and Training}
\label{sec:supp-method}

Pipeline overview and component diagram: main-paper Figure~1. Component structure, symbols and
losses are given in main-paper Section~4; this section records the sizes and optimization settings
behind them.

\subsection{Component Sizes}

The frozen Stage-1 stack holds $9.7$M parameters: $4.8$M in the context encoder $f_\theta$ (six
pre-norm transformer blocks, eight heads, width $256$, applied per day over the joint $50$-token
set) and $4.9$M in the block-causal autoregressive predictor $g_\phi$ (six blocks, eight heads,
width $256$, residual output with a zero-initialized projection). The target encoder $f_\xi$ is an
architectural copy of $f_\theta$ updated only by the exponential moving average, and the auxiliary
OLR reconstruction head is discarded after pretraining; neither is part of the released model.
Stage~2 holds $5.7$M parameters: a $2.1$M shared trunk (two token projections, a $3\times3$ scale
fusion at the $5\times5$ token grid, two transposed-convolution upsampling blocks to $20\times20$ at
$128$ channels, and a $1\times1$ OLR readout), $0.4$M in the rain-history and geography conditioning
encoders, a $3.0$M probability branch (U-Net width $96$, conditioning re-injected at every scale),
and a $0.3$M deterministic branch (the same U-Net topology at width $32$, without the noise and time
inputs); each component is rounded independently, so the parts sum to $5.8$M against the $5.7$M
total. Inference cost is itemized in Section~\ref{sec:supp-compute}.

The trunk and probability branch are trained jointly on main-paper Equation~(9); only the Stage-1
stack is frozen. The deterministic branch is trained afterward, as a separate readout on that same
trunk now held frozen, on main-paper Equation~(10); the two branches never share gradients into the
trunk. This sequencing is deliberate, because every deterministic-branch trunk trained end-to-end under the
deterministic loss alone tops out at pooled spatial ACC $0.255$--$0.271$ regardless of curriculum, while the
frozen, distributionally-trained trunk borrowed here reaches $0.292$ on the same pooled convention, consistent with a
distributional objective building more useful deterministic features than a deterministic loss does
on its own. Both branches' outputs are used directly, with no post-hoc recalibration. Retraining the
deterministic branch from three random seeds on the fixed frozen trunk (checked under this
project's earlier pooled scoring convention, since superseded by the case-averaged convention used
throughout the main text) moved the reported numbers within $\pm0.007$\,mm/day on RMSE, $\pm0.008$ on
spatial ACC, and $\pm0.010$--$0.020$ on FSS at the $1^\circ$--$5^\circ$ neighborhood widths, so
RMSE and ACC gains do not rest on a single seed. The released checkpoint is seed $1410$, which is
also the best of the three; we report the spread above rather than that seed's score alone, since
selecting and reporting the best of several seeds inflates apparent skill \citep{henderson2018}. The seed
spread is small enough that the tabulated values stand whichever seed is used, with one exception
noted below. Under the case-averaged convention, persistence outright leads
the deterministic branch's FSS at $3^\circ$ and $5^\circ$ neighborhood width (main-paper Section~5.2;
Table~\ref{tab:deterministic} below), superseding the earlier pooled-convention finding that this margin
was merely seed-sensitive. That FSS comparison is the exception noted above: the seed spread was
measured under the pooled convention and has not been re-measured under the case-averaged one, so we
treat the persistence-versus-M-JEPA neighborhood comparison as seed-sensitive and do not claim it.

\subsection{Optimization Settings}

Table~\ref{tab:supp-training} lists the settings for both stages. Stage~2 uses a fixed epoch budget
and the final-epoch checkpoint, with no validation-based selection, since on jointly trained trunks the
validation neighborhood score swings between evaluations, so best-checkpoint selection returns
half-trained models.

\begin{table}[h]
\centering\small
\begin{tabular}{@{}lll@{}}
\toprule
 & Stage~1 (pretraining) & Stage~2 (readout) \\
\midrule
Objective & Eq.~(4) & Eq.~(9) \\
Optimizer & AdamW & AdamW \\
Learning rate & $5\times10^{-4}$ & $10^{-3}$ \\
Weight decay & $0.05$ & $10^{-4}$ \\
Warmup & $10$ epochs & none \\
Epochs & $100$ & $80$ \\
Batch size & $32$ & $32$ \\
Gradient clipping & none & $1.0$ \\
Checkpoint & final epoch & final epoch \\
Augmentation & $\pm2$-cell jitter & $\pm2$-cell jitter \\
Rollout horizon & curriculum $2\rightarrow7$ & fixed at $7$ \\
VICReg (var/cov) & $1.0$ / $0.04$ & not applicable \\
Auxiliary weights & OLR $0.1$ & energy score $0.1$ \\
Trainable parameters & $9.7$M & $5.7$M \\
Windows & $3885$ (all season) & $667$ (rain-labeled) \\
Hardware & \multicolumn{2}{l}{single NVIDIA RTX A4000 (16\,GB)} \\
\bottomrule
\end{tabular}
\caption{Optimization settings for both training stages. Equation numbers refer to the main paper.
Stage~1 sees no rainfall at any point; Stage~2 trains only the trunk and the two branches, on the
frozen Stage-1 rollout.}
\label{tab:supp-training}
\end{table}

The controlled comparison of main-paper Table~3 uses the same flow-based probability head and the
same Stage-2 recipe on both frozen backbones, so that any skill difference reflects the pretraining
signal rather than the decoder. Because the rain-only control's decoder-training population ($226$
windows, against $667$) is small, its loader is given the same $\pm2$-cell spatial-jitter
augmentation used in the main pipeline. The three rain-history baselines have no frozen trunk to
attach a branch to and are trained end to end instead, with a standard zero-inflated
occurrence-plus-intensity loss on the architectures listed in Section~\ref{sec:supp-baselines}.

\subsection{Configuration Details for All Compared Systems}
\label{sec:supp-configs}

Table~\ref{tab:supp-configs} itemizes the architecture and training configuration for every model
compared in this paper. The three rain-history baselines (Section~\ref{sec:supp-baselines}) take a
five-day rainfall history as input
and predict occurrence and intensity jointly for all seven lead days; all three are trained with
AdamW ($\text{lr}=10^{-3}$, weight decay $10^{-4}$) for $60$ epochs on the identical tracked
patches and splits as the primary model. IFS, AIFS, and GraphCast are not trained by us; they
are public operational (IFS) or research (AIFS, GraphCast) forecast archives, cropped to the same
Lagrangian patch and lead as our own predictions and scored identically as single deterministic
members (Section~C).

\begin{table*}[t]
\centering\small
\begin{tabular}{@{}lp{6.7cm}p{6.7cm}@{}}
\toprule
Model & Architecture & Training \\
\midrule
M-JEPA & Multiscale transformer context encoder ($4.8$M) + autoregressive predictor ($4.9$M),
both frozen at readout; shared decoder trunk with two branches ($5.7$M) & Self-supervised on proxies
only, no rain objective; trunk and branches trained on paired rain data, $667$ train windows \\
Rain-only control & Identical architecture, backbone $9.6$M against M-JEPA's $9.7$M (only the
tokenizer's input-channel count differs) & Same recipe as M-JEPA, applied to rainfall alone; $226$
pretrain windows \\
U-Net & Encoder-decoder, 2 downsample / 2 upsample conv stages, skip connections, 32 base
channels (351K params) & AdamW, lr $10^{-3}$, weight decay $10^{-4}$, 60 epochs, rain-history
input only \\
ConvLSTM & Single ConvLSTM cell, 24 hidden channels, closed-loop autoregressive rollout (22K
params) & AdamW, lr $10^{-3}$, weight decay $10^{-4}$, 60 epochs, rain-history input only \\
ViT & Patch embedding (patch $4$, dim $128$), 4-layer/4-head transformer encoder (836K params) &
AdamW, lr $10^{-3}$, weight decay $10^{-4}$, 60 epochs, rain-history input only \\
IFS & Physics-based NWP, operational & Not trained by us; public open-data archive \\
AIFS & ML emulator \citep{aifs}; its ensemble variant is CRPS-trained \citep{aifscrps} & Not trained by us; public research archive \\
GraphCast & ML emulator, graph neural network \citep{graphcast} & Not trained by us; public
research archive \\
\bottomrule
\end{tabular}
\caption{Model and training configuration for every system compared in this paper. All
M-JEPA, rain-only-control, U-Net, ConvLSTM and ViT models are trained on the identical Lagrangian patches and
year-block splits (Section~A) so that comparisons isolate the modeling choice, not the data. IFS,
AIFS, and GraphCast are external references, evaluated but not trained.}
\label{tab:supp-configs}
\end{table*}

\section{Verification Protocol and Scoring Conventions}
\label{sec:supp-verification}

The main paper states these conventions once and then uses them without restatement.

\paragraph{Patch frame.} Every score is computed on the cells a tracked Lagrangian patch covers, not
on the full model domain. The operational archives are cropped to the same cells and the same lead
before scoring, so all systems are compared on identical geometry. Which cells count is fixed by an
\emph{observation-derived} patch mask: the valid-observation footprint of the target field defines
the scored cells, so the mask is set by the data rather than by any model's output, and no system can
be advantaged by the choice of scoring region. Full-domain rescoring of the archives, for contrast,
is in Section~\ref{sec:supp-acc-trend}.

\paragraph{Case-averaging.} Each (window, lead) pair is scored individually and the resulting scores
are averaged, rather than pooling every cell of every case into one score. Pooling lets the largest
and wettest cases dominate; case-averaging weights each forecast situation equally. The project
previously used the pooled convention, and where an older pooled number is quoted it is labeled as
such.

\paragraph{Probabilistic scores.} Brier skill is computed against a patch-local climatology at the
$1$\,mm threshold, following the recommendation that a climatological reference be local to the
verification sample rather than global. \emph{Member-FSS} applies the neighborhood fraction skill
score \citep{roberts2008} to each ensemble member separately and then averages, so it credits a
forecast that places rain within a spatial tolerance rather than requiring exact hits. FSS10 is that
score at the $10$\,mm threshold; the neighborhood width is stated with each table that reports it.

\paragraph{Anomaly correlation.} ACC $=\mathrm{corr}(\mathrm{pred}-\mathrm{clim},\,
\mathrm{obs}-\mathrm{clim})$ is scale-invariant: it rewards spatial pattern match and is blind to
amplitude, so a forecast that sheds fine-scale detail can score well precisely by being smooth. That
property drives the comparison in main-paper Section~5.2 and is analyzed in
Section~\ref{sec:supp-acc-trend}.

Matched-archive comparisons (M-JEPA vs.\ ECMWF 51-member) use the NHC (National Hurricane
Center) homogeneous-sample convention: two forecasts are paired and compared only on calendar days
where both forecasts have complete, valid output. This guards against archive-coverage biases.
In this study, $n{=}169$ days (2018--2025 test period) have both M-JEPA 7-lead and ECMWF
ensemble 7-lead output at the reference patch anchor. Windows whose system anchor lies near a
domain edge are retained and scored on their observed cells only, instead of being discarded by a naive
crop of the reference grid; this is why the matched population is not smaller than the archive
extent alone would suggest.

Fair CRPS estimator: we use the unbiased form $\text{CRPS}_\text{fair} = \frac{1}{K} \sum_i |x_i - y| - \frac{1}{2K(K{-}1)} \sum_{i,j} |x_i - x_j|$ \citep{ferro2008}, where $K$ is ensemble
size and $y$ is the observation. The biased ``energy'' form $\text{CRPS}_\text{energy} = \frac{1}{K} \sum_i |x_i - y| - \frac{1}{2K^2} \sum_{i,j} |x_i - x_j|$ systematically favors larger ensembles,
and comparison of unequal-member forecasts with the biased form can flip the sign of the result,
a pitfall relevant to any comparison of M-JEPA (16 members) vs.\ operational
ensembles (51 members).

\section{Extended Probing Results}
\label{sec:supp-probing}

This section gives the full per-lead numbers and extended analysis behind the probes in main-paper
Section~5: predictive information (Section~\ref{sec:supp-q1}), variable decodability and subspace
structure (Section~\ref{sec:supp-q2}), and CKA trajectory evolution (Section~\ref{sec:supp-q3}).
Section~\ref{sec:supp-q4} gives the latent-to-rainfall attribution analysis in full; for space, this
one is presented only here rather than summarized in the main paper.

\subsection{Predictive Information: Full Results by Lead}
\label{sec:supp-q1}

\begin{figure}[h]
\centering
\includegraphics[width=0.85\linewidth]{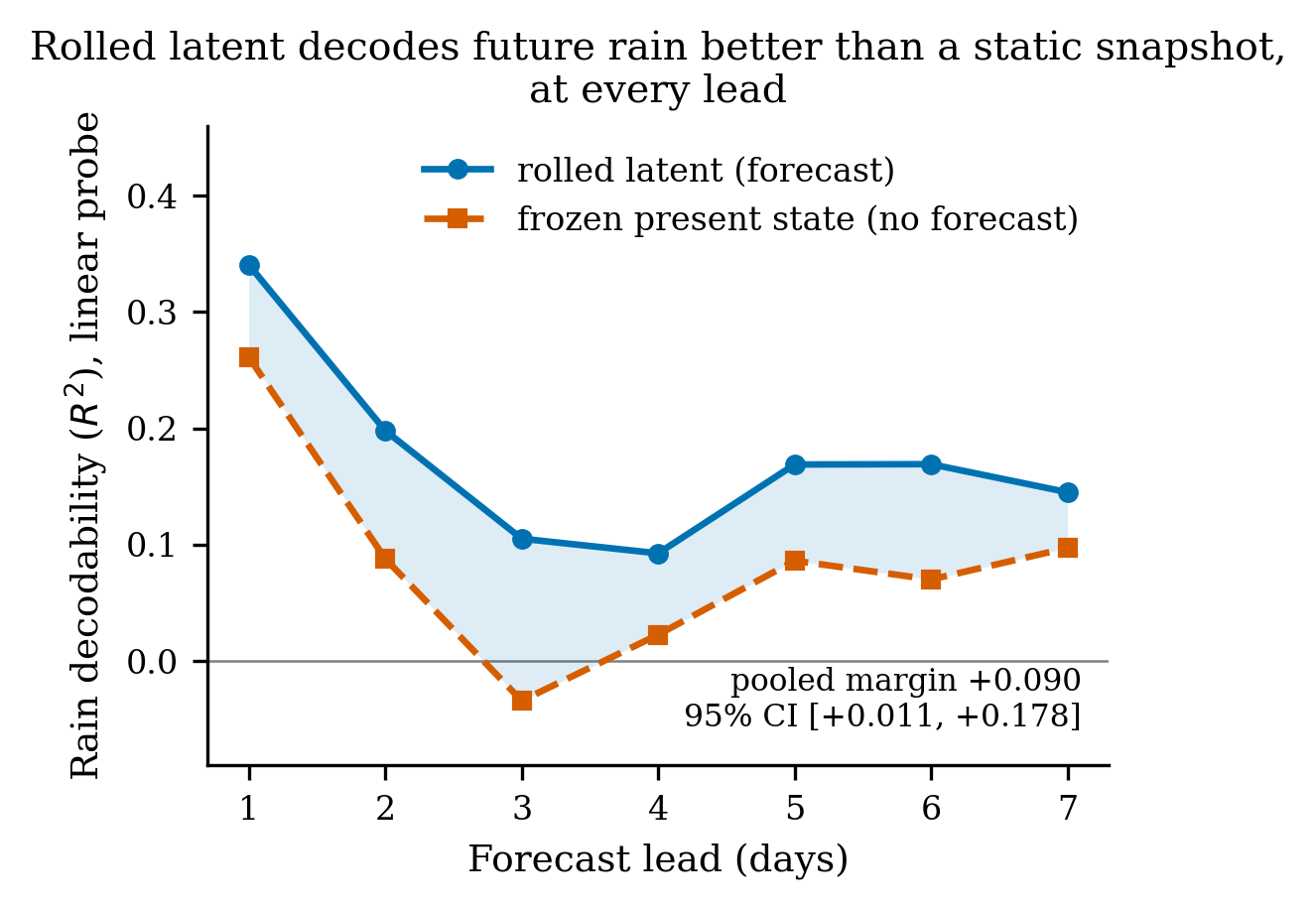}
\caption{Rain decodability ($R^2$, linear probe) at each lead from the rolled latent (forecast) vs.\
the frozen present-state control (no forecast). The rolled latent decodes future rainfall better at
every lead, by the widest margin at lead~3. Shaded region is the paired rolled-minus-frozen margin,
whose pooled value and $95\%$ bootstrap interval ($10^4$ resamples) are annotated; this paired margin,
not the two marginal curves separately, is the statistic the claim rests on.}
\label{fig:predinfo}
\end{figure}

Table~\ref{tab:supp-predinfo} gives per-lead R$^2$ for the rolled latent vs.\ the static-latent control (the
frozen context-end latent reused at every lead) across all 7 forecast days, and
Figure~\ref{fig:predinfo} plots the two curves with their paired margin. The margin is hump-shaped, largest at lead 3
(where the static probe turns negative, below the unconditional-mean baseline), smallest at lead 7
(rollout advantage erodes at long range). The negative R$^2$ for the static probe at lead 3
indicates the initial latent state underperforms the unconditional-mean predictor at that lead.

\begin{table}[h]
\centering\small
\resizebox{\columnwidth}{!}{%
\begin{tabular}{lcccc}
\toprule
Lead (days) & Rolled $R^2$ $\uparrow$ & Static $R^2$ $\uparrow$ & Margin $\uparrow$ & Margin 95\% CI \\
\midrule
1 & 0.341 & 0.261 & +0.079 & $[+0.004,+0.169]$ \\
2 & 0.198 & 0.088 & +0.110 & $[-0.003,+0.246]$ \\
3 & 0.105 & $-0.034$ & +0.139 & $[-0.003,+0.292]$ \\
4 & 0.092 & 0.022 & +0.070 & $[-0.076,+0.227]$ \\
5 & 0.169 & 0.086 & +0.083 & $[-0.051,+0.226]$ \\
6 & 0.169 & 0.070 & +0.099 & $[-0.048,+0.263]$ \\
7 & 0.145 & 0.097 & +0.048 & $[-0.096,+0.203]$ \\
\midrule
Pooled & \textbf{0.174} & 0.084 & \textbf{+0.090} & $\mathbf{[+0.011,+0.178]}$ \\
\bottomrule
\end{tabular}%
}
\caption{Rolled vs.\ static-latent R$^2$ per lead. Margin peaks at lead 3, consistent with
a forecast whose advantage over persistence erodes at long range. The $95\%$ confidence interval is
from a $10^4$-replicate paired bootstrap over test windows; the pooled (lead-averaged) margin
excludes zero, while individual leads overlap zero given the modest test-window count.}
\label{tab:supp-predinfo}
\end{table}

\subsection{Variable Decodability and Subspace Structure: Extended Analysis}
\label{sec:supp-q2}

Table~\ref{tab:supp-probes} reports linear-probe and CV-tuned nonlinear-probe R$^2$ for the five proxy
variables. The nonlinear probe is a multilayer perceptron whose
architecture and optimizer are chosen by a cross-validated grid search over hidden width
$\{128,256,512\}$, depth $\{1,2\}$, learning rate $\{10^{-3},3\times10^{-4}\}$, and weight decay
$\{10^{-3},10^{-4}\}$ (24 configurations), with model selection on the validation years
(2014--2017) only; the reported
nonlinear R$^2$ is the selected configuration's \emph{test}-year score, and the test years are never
used for selection. The validation years do fall inside the self-supervised pretraining span, so
probe selection is not fully independent of the encoder's training data. That can only favor the
tuned nonlinear probe, and the finding here is that the tuned probe does not exceed the linear one,
so the contamination works against the conclusion we draw rather than for it. The proxies are strongly linearly decodable (0.67--0.85), and the tuned
nonlinear probe does not significantly outperform the linear probe on any of them, and the paired-bootstrap gap
confidence interval includes zero for four variables and is significantly \emph{negative} for
humidity, with the largest point gain only $+0.02$ (geopotential, OLR envelope). That the tuned probe adds nothing indicates the variable content is stored in a linearly accessible
form rather than a hidden nonlinear code.

\begin{table}[h]
\centering\small
\resizebox{\columnwidth}{!}{%
\begin{tabular}{lccc}
\toprule
Latent Dimension & Linear $R^2$ $\uparrow$ & CV-tuned MLP $R^2$ $\uparrow$ & Gap [95\% CI] \\
\midrule
$q$ (humidity) & 0.697 & 0.644 & $-0.052\ [-0.105,-0.007]$ \\
$u$ (zonal wind) & 0.668 & 0.642 & $-0.026\ [-0.082,+0.029]$ \\
$v$ (merid.\ wind) & 0.679 & 0.668 & $-0.011\ [-0.062,+0.039]$ \\
$z$ (geopotential) & 0.849 & 0.865 & $+0.016\ [-0.005,+0.037]$ \\
OLR (smoothed) & 0.744 & 0.763 & $+0.018\ [-0.021,+0.060]$ \\
\bottomrule
\end{tabular}%
}
\caption{Linear vs.\ CV-tuned nonlinear probe R$^2$ for the five proxy readout targets, with the
paired-bootstrap ($10^4$-replicate) $95\%$ CI on the tuned-minus-linear gap. Proxies are strongly and
linearly decodable, and the tuned MLP does not significantly exceed the linear probe on any of them.
Selected MLP configurations: $q,u$ width $512$/depth
$1$; $v$ width $128$/depth $1$; $z$ width $256$/depth $1$; OLR width $256$/depth $2$.}
\label{tab:supp-probes}
\end{table}

The five proxy directions occupy near-orthogonal subspaces (mean absolute off-diagonal cosine
similarity 0.15 among the proxies alone). Overlaps are physically interpretable, with geopotential and the
OLR envelope couple at $+0.38$, humidity and the meridional wind at $+0.30$, and the two horizontal
wind components at $-0.20$: deep convection organized beneath lowered geopotential, and moisture
advected by the northward monsoon flow.

\subsection{CKA Evolution and Latent-Increment Structure}
\label{sec:supp-q3}

Centered Kernel Alignment (CKA) between the context latent and the rolled latent at each of the seven
rollout steps shows the trajectory's saturation (Table~\ref{tab:supp-cka}), with a rapid fall from
0.944 at lead 1 to 0.794 by lead 5,
then plateau. This
is an independent signature of damped intraseasonal dynamics in the learned representation.

\begin{table}[h]
\centering\small
\begin{tabular}{lc}
\toprule
Lead (days) & CKA (context vs.\ rollout) \\
\midrule
1 & 0.944 \\
2 & 0.872 \\
3 & 0.825 \\
4 & 0.804 \\
5 & 0.794 \\
6 & 0.793 \\
7 & 0.794 \\
\bottomrule
\end{tabular}
\caption{CKA between the context latent and the rolled latent at each of the seven rollout steps
(leads~1--7). Rapid initial drop (evolution), then plateau (saturation).}
\label{tab:supp-cka}
\end{table}

Principal-component analysis of latent increment vectors $\hat z_{t+d} - z_t$ reveals distributed
evolution: the top-3 PCs explain only 16\%, 15\%, 11\% of increment variance respectively, totaling
42\%, with no single dominant axis of change. This suggests the rollout evolves through many
weakly-coupled modes, consistent with an environment interacting with the tracked system (wind shear,
moisture convergence) rather than a single low-dimensional mode of variation.

\subsection{Attribution and Saliency: Latent-Space Gradients}
\label{sec:supp-q4}

\begin{figure}[t]
\centering
\includegraphics[width=\linewidth]{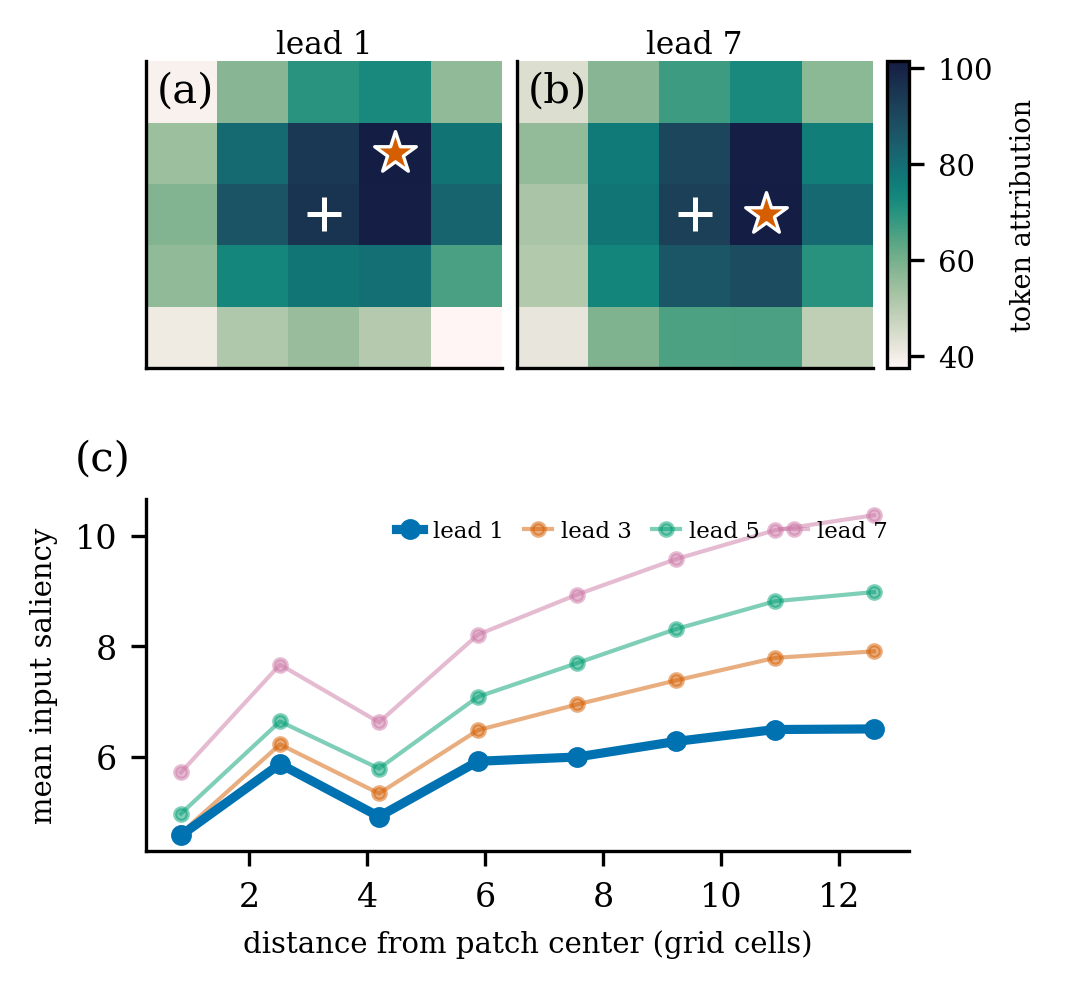}
\caption{The readout draws on tokens around, not exactly on, the tracked convective centroid.
(\textbf{a},\textbf{b}) Token-level attribution (gradient of expected rainfall with respect to the
rolled foreground latent tokens, native $5\times5$ token grid, no smoothing, arbitrary units, shared
color scale) at leads~1 and~7; the star marks the peak token and the cross the patch's geometric
center; north is up and east is right. The peak sits one cell north-east of center at lead~1 and one
cell east of center at lead~7. (\textbf{c}) Radially binned mean input-space saliency (gradient with
respect to the raw $20\times20$ foreground field) against distance from the patch center, per lead:
increases toward the patch edge rather than peaking at a fixed ring, a pattern we attribute to the
decoder's convolutional boundary handling rather than to convective structure.}
\label{fig:probe_attr}
\end{figure}

We trace how latent content becomes a rainfall prediction, on the current production decoder
(shared flow trunk plus the det-on-flow-trunk deterministic head; an earlier version of this
analysis used a since-retired decoder head and is superseded by the numbers below). We backpropagate
the decoded expected-rain field to the rolled foreground latent tokens and aggregate the gradient
magnitude per token on the native $5\times5$ token grid. The driving token sits one cell off the
patch center (north and east of it) and is stable across leads~1--6, shifting to directly east of
center at the longest lead (Figure~\ref{fig:probe_attr}a,b); the center token itself is high but
never the maximum ($91$--$99\%$ of the peak across all seven leads). The grid's contrast falls from a
coefficient of variation of $0.272$ at lead~1 to $0.240$ at lead~7 (a $\approx12\%$ reduction, not
fully monotonic), consistent with the mild dispersion expected from the trajectory saturation
reported in Section~\ref{sec:supp-q3}. A complementary input-space saliency map at pixel resolution
increases with distance from the patch center rather than peaking at a fixed ring
(Figure~\ref{fig:probe_attr}c); this pattern is architecture-dependent and consistent with a
convolutional boundary effect in the decoder trunk rather than necessarily reflecting the true
convective footprint, so we do not read a specific radius from it. The token-level result is the
more direct evidence, since the pretrained latent drives the readout through an off-center, spatially
stable set of tokens rather than the exact centroid token alone.

\section{Controls and Ablations}
\label{sec:supp-controls}

Main-paper Section~5.6 isolates the source of the transfer with two controls (rainfall-only
pretraining and a randomly initialized backbone), a further generalization test on a second held-out
target, a leave-one-proxy-out ablation, and a check of sensitivity to the tracker's detection
settings, which establishes that the
object-based evaluation frame itself is not an artifact of a particular choice of detection
threshold or minimum object area. Each is given in full below, in that order.

\subsection{Rainfall-Only Pretraining and Rain-History Baselines}
\label{sec:supp-baselines}

To isolate the value of proxy-based pretraining, we retrain the identical JEPA encoder and predictor
using the same self-supervised recipe on rainfall alone, holding architecture and training procedure
constant while replacing the five proxy fields with rainfall only. This tests whether the
proxy-learning hypothesis is driven by proxy availability (information-limited) or by something about
the architecture or loss function (idea-limited).

The main paper's claim that this rain-only control is stronger than conventional rain-history
architectures requires ruling out that such an architecture, given only rainfall history, could do
better if better designed or better parameterized. We therefore trained three standard architectures
on the identical tracked patches and splits, each taking only a five-day rainfall history as input,
with a standard zero-inflated occurrence-plus-intensity loss: a compact U-Net (351K parameters), a
ConvLSTM (22K parameters), and a Vision Transformer (836K parameters).
Table~\ref{tab:supp-rainbaselines} gives the result, and expands main-paper Table~3 with the
neighborhood-skill column.

\begin{table}[h]
\centering\small
\resizebox{\columnwidth}{!}{%
\begin{tabular}{lrccc}
\toprule
Model & Params & CRPS (mm/day) $\downarrow$ & FSS10 $\uparrow$ & BSS $\uparrow$ \\
\midrule
M-JEPA (proxies) & 9.7M & \textbf{5.54} & 0.53 & \textbf{0.18} \\
Rain-only control (same recipe) & 9.6M & 7.52 & 0.54 & $-0.04$ \\
U-Net (rain only) & 351K & 7.38 & 0.14 & $-0.62$ \\
ConvLSTM (rain only) & 22K & 8.33 & 0.00 & $-1.07$ \\
ViT (rain only) & 836K & 8.65 & 0.42 & $-0.99$ \\
\bottomrule
\end{tabular}%
}
\caption{Rain-history baselines against the primary model and the same-recipe rain-only control, on
the same tracked patches and year-block splits, case-averaged and without recalibration. Parameter
counts are the self-supervised backbone for the two JEPA rows
(context encoder plus predictor) and the full trained network for the three baselines. All rows are
scored on the full test set ($n{=}337$ windows) except the rain-only control, whose sparse rain-valid
data leave it a smaller held-out population ($n{=}118$), a consequence of the very sparsity the control is designed to
expose. The three
standard architectures span two orders of magnitude in parameter count (22K--836K) yet all have
\emph{negative} Brier skill, below the climatological reference at the 1\,mm threshold, so their
failure is one of information, not capacity. The rain-only control's Brier skill is closest to zero
among the rain-only models, which supports the main-paper claim that the self-supervised recipe
contributes substantively once given adequate information.}
\label{tab:supp-rainbaselines}
\end{table}

Under-parameterization is not the cause of the standard architectures' failure: they fail
identically across two orders of magnitude in size, so rainfall history alone is an insufficient
basis for this forecasting task at this lead range. The rain-only control's self-supervised objective,
though itself limited by the sparse target signal, still extracts more from that target than any of
the three conventional networks.

The failure is invariant to capacity from two directions, which rules out the proxy model
outperforming merely because it is larger. Downward, the three rain-history
baselines fail identically across $22$K--$836$K parameters, so adding parameters to a rain-only model
does not help. Upward, the control's backbone holds $9.6$M parameters against M-JEPA's $9.7$M on the
identical architecture and self-supervised recipe, the input-channel count of the tokenizer being the
only difference, and it still trails the proxy model substantially (CRPS $7.52$ against
$5.54$\,mm/day, Brier skill $-0.04$ against $+0.18$). Matching the parameter budget while holding the
input to rainfall alone therefore does not eliminate the difference. What changes performance is the
pretraining \emph{signal} rather than the parameter count, which the control is designed to
isolate. This graceful, information-tracking degradation is also the basis for the scaling hypothesis
of Section~\ref{sec:supp-scaling}.

\subsection{Randomly Initialized Backbone}
\label{sec:supp-pt1}

A second control isolates \emph{learning} from \emph{architecture}: per-lead R$^2$ for the trained
encoder vs.\ a randomly initialized backbone (same multiscale transformer + recurrent rollout
architecture, no pretraining), in Table~\ref{tab:supp-randinit} and
Figure~\ref{fig:supp-probe-controls}a. The trained model separates at every lead; random skill
is residual ($\approx 0.10$ by leads 6--7), reflecting target smoothness and not structure
inherent to the untrained architecture. Ratios are computed from the unrounded per-lead values.
The trained column comes from a separate execution of the same probe on the same checkpoint and the
same $667$/$337$ window split as Table~\ref{tab:supp-predinfo}; the probe fit is stochastic, so the
per-lead values differ from that table by up to $0.037$ and the pooled value by $0.003$ ($0.177$
against $0.174$). We report each run against its own paired control rather than mixing the two.

\begin{table}[h]
\centering\small
\resizebox{\columnwidth}{!}{%
\begin{tabular}{lccccc}
\toprule
Lead (days) & Trained $R^2$ $\uparrow$ & Random $R^2$ $\uparrow$ & Ratio $\uparrow$ & Gap $\uparrow$ & Gap 95\% CI \\
\midrule
1 & 0.336 & 0.040 & 8.4× & +0.296 & $[+0.183,+0.415]$ \\
2 & 0.189 & 0.004 & 42.7× & +0.185 & $[+0.078,+0.294]$ \\
3 & 0.086 & 0.018 & 4.9× & +0.068 & $[-0.060,+0.200]$ \\
4 & 0.124 & 0.024 & 5.3× & +0.101 & $[-0.018,+0.218]$ \\
5 & 0.132 & 0.048 & 2.7× & +0.083 & $[-0.037,+0.206]$ \\
6 & 0.187 & 0.102 & 1.8× & +0.085 & $[-0.030,+0.205]$ \\
7 & 0.182 & 0.104 & 1.8× & +0.078 & $[-0.039,+0.202]$ \\
\midrule
Pooled & \textbf{0.177} & 0.049 & \textbf{3.6×} & \textbf{+0.128} & $\mathbf{[+0.055,+0.205]}$ \\
\bottomrule
\end{tabular}%
}
\caption{Trained vs.\ random-init backbone R$^2$. The trained column is the same rolled-latent
predictive-information probe reported in main-paper Section~5.5, re-run alongside its random-init
control (see text for the run-to-run spread). Learning
multiplies decodability by 5--43× over leads 1--4, converging to $\approx$1.8× at the longest leads
as random-feature decodability of a smooth, seasonally autocorrelated target rises; residual random
skill is not attributed to the representation. The gap $95\%$ CI is a $10^4$-replicate paired
bootstrap over test windows computed under the original pairing; the pooled gap and the near-term
leads exclude zero.}
\label{tab:supp-randinit}
\end{table}

\subsection{Cross-Target Transfer (OLR)}
\label{sec:supp-pt2}

\begin{figure}[t]
\centering
\includegraphics[width=0.95\linewidth]{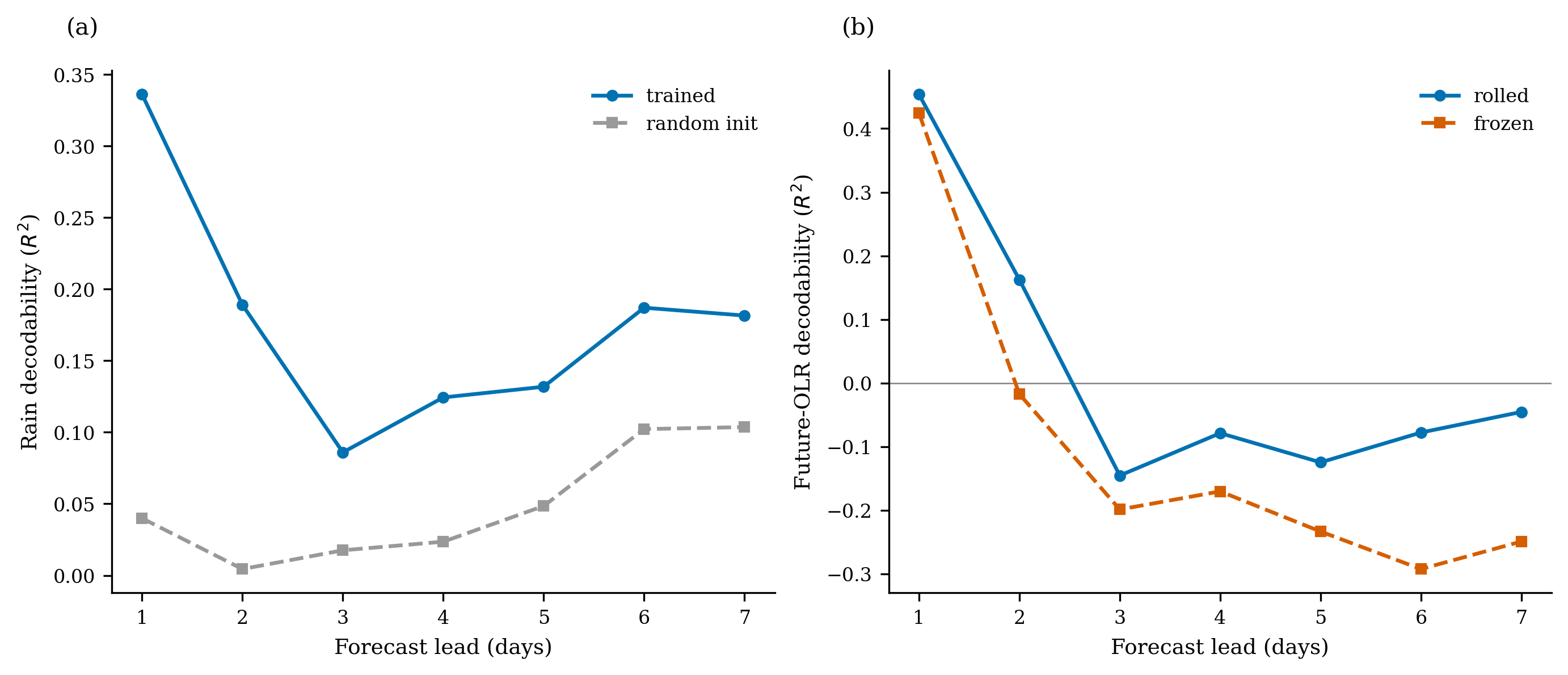}
\caption{Controls isolating the source of transfer (main-paper Section~5.6), shown visually.
(\textbf{a}) Trained encoder vs.\ randomly initialized backbone of identical architecture: both tested
on the same predictive-information probe (rolled-vs-static R$^2$). The trained model separates from
the random one at every lead; random features retain residual long-lead decodability
($\approx0.10$ by leads~6--7, indicating target smoothness), while trained skill stays at least
$1.4\times$ higher at every lead, confirming the transfer is a product of learning. (\textbf{b}) Cross-target transfer: rolled-vs-static
R$^2$ margin for a held-out second target (future OLR anomaly, never seen during rain-decoder
training). Margin larger than rainfall ($+0.126$ vs.\ $+0.090$) and broadens with lead ($+0.03$
near-term, $+0.20$ long-range), though absolute skill drops negative at long leads. Demonstrates the
pretrained representation encodes general atmospheric state, not rain-specific shortcuts.}
\label{fig:supp-probe-controls}
\end{figure}

If the pretrained latent encodes general atmospheric state, its predictive advantage should not be
specific to rainfall. Table~\ref{tab:supp-crosstarget} and Figure~\ref{fig:supp-probe-controls}b give
rolled vs.\ static-latent R$^2$ for the future-OLR readout at each lead. The
margin is non-monotonic:
small at lead 1 ($+0.03$), largest at leads 6--7 ($\approx +0.21$). Absolute skill is low: the
rolled probe is positive only at leads 1--2 and negative beyond, and the static probe is negative
from lead 2 onward, since forecasting the OLR anomaly that far ahead from a single latent is
difficult. The quantity of interest is therefore the \emph{relative} advantage of the rollout
over the static latent, which is positive at every lead and widens as both probes decay; this is
what shows the learned state transfers to a second, held-out atmospheric field.

\begin{table}[h]
\centering\small
\resizebox{\columnwidth}{!}{%
\begin{tabular}{lcccc}
\toprule
Lead (days) & Rolled $R^2$ $\uparrow$ & Static $R^2$ $\uparrow$ & Margin $\uparrow$ & Margin 95\% CI \\
\midrule
1 & 0.454 & 0.424 & +0.030 & $[-0.058,+0.121]$ \\
2 & 0.163 & $-0.017$ & +0.180 & $[+0.047,+0.322]$ \\
3 & $-0.145$ & $-0.199$ & +0.053 & $[-0.124,+0.226]$ \\
4 & $-0.079$ & $-0.170$ & +0.090 & $[-0.081,+0.256]$ \\
5 & $-0.125$ & $-0.234$ & +0.110 & $[-0.096,+0.298]$ \\
6 & $-0.078$ & $-0.293$ & +0.215 & $[-0.001,+0.419]$ \\
7 & $-0.045$ & $-0.249$ & +0.204 & $[-0.009,+0.398]$ \\
\midrule
Pooled & 0.021 & $-0.105$ & \textbf{+0.126} & $\mathbf{[+0.016,+0.233]}$ \\
\bottomrule
\end{tabular}%
}
\caption{Rolled vs.\ static-latent R$^2$ for future-OLR anomaly (cross-target transfer). Margin positive
at every lead; absolute skill negative beyond lead 2, but the relative advantage of the rollout
persists and widens. The $95\%$ CI is a $10^4$-replicate paired bootstrap over test windows; the
pooled margin excludes zero.}
\label{tab:supp-crosstarget}
\end{table}

The widening margin is consistent with the CKA trajectory's lead-5 plateau
(Section~\ref{sec:supp-q3}), despite an apparent tension between a representation whose structure
saturates by lead 5 and a cross-target margin that continues to widen through leads 6--7. The CKA
trajectory is computed once on the latent rollout itself (context latent versus rolled latent at
each lead), independent of which downstream target (rain or future OLR) reads it out, so the lead-5
saturation is a property of the latent dynamics, not of the rain decoder. Decomposing the margin
into its two arms resolves the apparent tension: the widening does not come from the rolled probe
gaining skill (its R$^2$ stays approximately flat and low across leads 3--7) but from the static-latent
control \emph{collapsing}, from $-0.017$ at lead~2 to $-0.293$ at lead~6, as a single frozen snapshot
becomes an increasingly poor predictor of the more distant OLR field. The growing margin therefore measures the decay of the
persistence baseline, not growth of rollout information, and is fully consistent with the lead-5 CKA
plateau: the rollout saturates while the static control keeps degrading.

\subsection{Leave-One-Proxy-Out Ablation}
\label{sec:supp-loo}

The controls above show that pretraining on the five proxies jointly transfers better than
pretraining on rainfall alone, but not that information from one proxy specifically reaches the
rainfall readout through the shared latent, as opposed to the five proxies simply supplying more
raw pretraining volume. We test this directly: retrain the full pipeline with one background
variable ($q$, $u$, $v$, or $z$) held out of the proxy set, everything else unchanged, and score
each ablated model's probability-branch forecast the same way as the full model (pooled CRPS, BSS,
member-FSS10; matched full test population), in Table~\ref{tab:supp-loo}. None of $q$, $u$, $v$, $z$ is itself the readout
target, so a drop in rainfall skill when one is removed is evidence that its information
was being combined with the others in the latent and used for rain prediction, not merely padding
the pretraining sample count.

\begin{table}[h]
\centering\small
\begin{tabular}{lccc}
\toprule
Proxy set & CRPS $\downarrow$ & BSS $\uparrow$ & FSS10 $\uparrow$ \\
\midrule
All five (full model) & \textbf{5.78} & \textbf{0.18} & 0.60 \\
$-$ humidity ($q$) & 5.80 & 0.17 & 0.58 \\
$-$ zonal wind ($u$) & 5.81 & 0.17 & 0.59 \\
$-$ meridional wind ($v$) & 5.88 & 0.16 & 0.56 \\
$-$ geopotential ($z$) & 5.81 & \textbf{0.18} & \textbf{0.60} \\
\bottomrule
\end{tabular}
\caption{Leave-one-proxy-out ablation, pooled convention, full test population. These runs predate
the switch to case-averaged scoring and are internally comparable to each other, not to the
case-averaged $5.54$ of main-paper Table~3. Every removal costs
CRPS, but the margins are modest ($0.5$--$1.8\%$) rather than large. Percentages here and in the
text are computed from the unrounded scores, not from the two-decimal entries shown, which round to
slightly different ratios. Meridional wind is the one
consistently load-bearing variable, the only ablation that also costs BSS and FSS10 together;
geopotential's removal leaves BSS and FSS10 essentially unchanged ($-1.0\%$ and $+0.1\%$).}
\label{tab:supp-loo}
\end{table}

Every ablation costs CRPS, so every proxy contributes some non-redundant information, but the
margins are small next to the $36\%$ CRPS gap between the five-proxy and rain-only models
(main-paper Table~3): no single proxy is doing the work of the whole set, consistent with
redundant, distributed encoding across the five rather than one dominant channel. Meridional wind is
the exception: its removal is the only one that degrades all three metrics together ($+1.8\%$
CRPS, $-12.8\%$ BSS, $-6.1\%$ FSS10), consistent with the humidity--meridional-wind coupling already
identified in the latent's subspace structure (main-paper Section~5.4): the moisture
transport that organizes convection is carried jointly by the two, so removing either weakens that
signal. Geopotential's removal is the mildest on BSS and FSS10, which it leaves unchanged, though humidity
costs marginally less on CRPS ($+0.50\%$ against $+0.63\%$); either way geopotential's contribution
is more redundant with the other four than load-bearing on its own. OLR is not ablated, because it
defines the tracking frame: removing it would change which windows exist, confounding an input
ablation with a change of evaluation population.

\subsection{Tracker and Detection-Setting Sensitivity}
\label{sec:supp-tracker}

A recurring concern with any object-based, tracker-defined evaluation frame is that the reported
ranking could be an artifact of the particular detection settings rather than a property of the
model \citep{prein2024}. We test this directly with a sensitivity grid over the two tracker
parameters that fix patch geometry, detection quantile $\{8,10,12\}\%$ and minimum object area
$\{0.75,1,1.5\}\times$ the default, re-tracking the held-out test years and re-scoring the frozen
model and a dressed-persistence ensemble on each of the nine resulting window populations plus the
default control (a configuration-grid sweep with window-bootstrap intervals, following the
cell-tracking sensitivity protocol of \citet{celltao}). The model-versus-baseline ranking
\emph{direction} is invariant across all ten settings: the model outperforms persistence on CRPS and
Brier skill in every configuration and outperforms the deterministic IFS in the same direction on
every matched subset, while the wet-cell fraction spans only $0.014$ across an order-of-magnitude
change in the detection threshold. These comparative conclusions are thus robust to, not
contingent on, the tracker choice. The sweep covers the persistence and deterministic-IFS
comparisons only; re-running the $51$-member ensemble comparison on all ten window populations was
beyond the archive access available to us, so the headline probabilistic result of
Section~\ref{sec:supp-readout} rests on the default configuration alone.

\section{Forecast Skill: Extended Results}
\label{sec:supp-readout}

Main-paper Table~1 gives the full skill matrix (CRPS, BSS, FSS10, Spread/RMSE, rank
histogram deviation) for M-JEPA Ens (16 members), IFS Ens (51 members), and the IFS deterministic
control, on matched windows ($n{=}169$, homogeneous-sample rule, fair unbiased CRPS estimator,
paired bootstrap with $10^4$ resamples); not repeated here. Rank histogram deviation is the mean
absolute departure from uniform across bins, lower better. An anomaly-correlation number is not
reported for this $n{=}169$ population, since it was not computed on this matched set; the
deterministic anomaly-correlation comparison against IFS, AIFS, and GraphCast on their own larger
matched populations is reported separately below and favors the operational references.

\subsection{Deterministic Skill by Lead}
\label{sec:supp-deterministic}

\begin{table}[t]
\centering\small
\caption{Deterministic skill on the full test set ($n{=}337$), case-averaged over the seven-day lead
window (each window/lead scored individually, then averaged; ACC: spatial anomaly correlation; FSS
at the $10$\,mm threshold \citep{roberts2008}), M-JEPA scored with the deterministic branch.
\emph{Ensemble mean} averages the probability branch's $16$ raw members into a point forecast, scored
identically, to check whether a dedicated point-forecast head earns its keep over simply averaging
the generative samples. Boldface marks the best value. Matched-subset comparison against IFS, AIFS,
and GraphCast: Table~\ref{tab:supp-crossmodel-summary}; per-lead detail below.}
\label{tab:deterministic}
\begin{tabular}{lcccc}
\toprule
System & RMSE$\,\downarrow$ & ACC$\,\uparrow$ & FSS$1^\circ$ & FSS$3/5^\circ$ \\
\midrule
M-JEPA & \textbf{16.24} & \textbf{0.31} & \textbf{0.35} & 0.47/0.53 \\
Ensemble mean & 16.48 & 0.29 & 0.34 & 0.48/0.55 \\
Persistence & 23.46 & 0.18 & 0.33 & \textbf{0.49}/\textbf{0.56} \\
\bottomrule
\end{tabular}
\end{table}

Dry-area specificity by lead is tabulated in main-paper Table~2, which keeps each reference on its
own matched subset and therefore reports two M-JEPA rows; we do not additionally plot it, since
collapsing those subsets into a single curve would average over non-homogeneous populations and
violate the sampling rule of Section~\ref{sec:supp-verification}.
Table~\ref{tab:supp-deterministic} gives the
full per-lead breakdown of the deterministic comparison summarized in the main paper (RMSE, spatial
anomaly correlation), for M-JEPA's deterministic branch and the ensemble-mean point forecast (the
probability branch's $16$ raw members averaged, Table~\ref{tab:deterministic}'s ensemble-mean row) against
persistence, case-averaged over the full $n{=}337$-window test set for each lead, true patch frame
throughout.

\begin{table}[h]
\centering\small
\begin{tabular}{lccc}
\toprule
Lead (days) & M-JEPA (det) & Ensemble mean & Persistence \\
\midrule
\multicolumn{4}{l}{\emph{RMSE (mm/day) $\downarrow$}} \\
1 & 16.79 & 17.20 & 21.12 \\
2 & 17.13 & 17.46 & 23.97 \\
3 & 16.88 & 17.12 & 24.30 \\
4 & 16.39 & 16.62 & 23.96 \\
5 & 15.99 & 16.13 & 23.73 \\
6 & 15.51 & 15.68 & 23.69 \\
7 & 14.97 & 15.13 & 23.45 \\
\midrule
\multicolumn{4}{l}{\emph{Spatial ACC $\uparrow$}} \\
1 & 0.447 & 0.409 & 0.384 \\
2 & 0.364 & 0.328 & 0.201 \\
3 & 0.313 & 0.291 & 0.148 \\
4 & 0.279 & 0.259 & 0.139 \\
5 & 0.266 & 0.256 & 0.129 \\
6 & 0.273 & 0.257 & 0.122 \\
7 & 0.261 & 0.244 & 0.114 \\
\midrule
\multicolumn{4}{l}{\emph{FSS10 at $5^\circ$ neighborhood $\uparrow$}} \\
1 & 0.721 & 0.726 & 0.756 \\
2 & 0.633 & 0.641 & 0.606 \\
3 & 0.560 & 0.573 & 0.548 \\
4 & 0.493 & 0.513 & 0.536 \\
5 & 0.451 & 0.486 & 0.513 \\
6 & 0.438 & 0.476 & 0.493 \\
7 & 0.414 & 0.450 & 0.473 \\
\bottomrule
\end{tabular}
\caption{Per-lead breakdown of Table~\ref{tab:deterministic}'s comparison against persistence, same
convention and routines throughout ($10$\,mm threshold, $5^\circ$ neighborhood). Persistence is
rebuilt here as the last observed context field at the fixed anchor; the lead-averaged values
reproduce Table~\ref{tab:deterministic}'s upper block exactly.}
\label{tab:supp-deterministic}
\end{table}

The same sharpness trade-off that shapes the probabilistic comparison (main-paper Table~1), where
M-JEPA accepts a neighborhood-skill cost relative to the operational ensemble in exchange for a CRPS
advantage, appears again here against persistence: the deterministic branch is uniformly better on
point accuracy and pattern correlation, at $7$--$8$\,mm/day lower RMSE and roughly double the spatial
ACC from lead~2 onward, while persistence retains a small advantage at the $5^\circ$ neighborhood
at lead~1 and from lead~4 onward. A field that relays the last observed rain distribution scores well on coarse-scale
overlap and poorly on placement and amplitude, which is what the two metric families report here.
Persistence being uncompetitive on RMSE and ACC at every lead confirms the rollout is not simply
relaying the initial condition forward.

The matched-coverage comparison against IFS, AIFS, and GraphCast summarized in main-paper
Section~5.2 (Table~\ref{tab:supp-crossmodel-summary} below) restricts M-JEPA to the cells its
tracked patches cover: the same
patch-native frame as main-paper Table~1, not the full $40\times40$ domain the references forecast
natively. Each system is compared on its own matched-coverage subset ($n=266$, $127$, $127$; the
all-five intersection is only $\approx14$ days, so the larger pairwise subsets are used and
disclosed). Each lead is scored independently over every valid tracked window reaching that
(calendar day, lead) pair, standard object-verification practice (e.g.\ tropical-cyclone track
verification, MODE), not a requirement that one single tracked window supply all seven leads.

\begin{table}[h]
\centering\small
\caption{Deterministic skill, case-averaged over the seven-day lead window, M-JEPA re-scored on each
reference's own matched-coverage subset (IFS: $n{=}266$; AIFS and GraphCast: $n{=}127$, an identical
subset by construction, reported once). \emph{Ensemble mean} as in Table~\ref{tab:deterministic}.
Every entry is the mean of the corresponding per-lead column of Table~\ref{tab:supp-crossmodel},
each lead carrying equal window count, so the two tables reconcile term by term. The anomaly
correlation uses a per-lead climatology throughout, as elsewhere in this paper; a climatology pooled
across leads raises every ACC in this table by $0.005$--$0.010$ without changing any ordering.}
\label{tab:supp-crossmodel-summary}
\begin{tabular}{lcccc}
\toprule
System & RMSE$\,\downarrow$ & ACC$\,\uparrow$ & FSS$1^\circ$ & FSS$3/5^\circ$ \\
\midrule
\multicolumn{5}{l}{\emph{IFS matched subset ($n{=}266$); M-JEPA re-scored}} \\
M-JEPA & \textbf{18.23} & 0.283 & 0.34 & 0.46/0.52 \\
Ensemble mean & 18.50 & 0.255 & 0.33 & 0.47/0.54 \\
IFS & 18.45 & 0.328 & 0.44 & 0.60/0.67 \\
\midrule
\multicolumn{5}{l}{\emph{AIFS/GraphCast matched subset ($n{=}127$); M-JEPA re-scored}} \\
M-JEPA & 19.74 & 0.339 & 0.37 & 0.49/0.56 \\
Ensemble mean & 20.14 & 0.294 & 0.35 & 0.49/0.56 \\
AIFS & 18.92 & 0.419 & 0.49 & 0.63/0.69 \\
GraphCast & 18.81 & 0.418 & 0.46 & 0.60/0.65 \\
\bottomrule
\end{tabular}
\end{table}

\begin{table*}[t]
\centering\small
\resizebox{\textwidth}{!}{%
\begin{tabular}{lccccccccc}
\toprule
Lead & \multicolumn{3}{c}{RMSE $\downarrow$} & \multicolumn{3}{c}{ACC $\uparrow$} & \multicolumn{3}{c}{FSS $\uparrow$} \\
 & det. & ens. & ref & det. & ens. & ref & det. & ens. & ref \\
\midrule
\multicolumn{10}{l}{\emph{vs.\ IFS ($n{=}161$)}} \\
1 & 19.03 & 19.49 & 20.18 & 0.410 & 0.366 & 0.367 & 0.725 & 0.723 & 0.759 \\
2 & 19.27 & 19.64 & 19.49 & 0.326 & 0.284 & 0.355 & 0.624 & 0.636 & 0.717 \\
3 & 18.97 & 19.19 & 18.98 & 0.282 & 0.259 & 0.349 & 0.535 & 0.555 & 0.695 \\
4 & 18.41 & 18.72 & 18.57 & 0.248 & 0.220 & 0.319 & 0.481 & 0.492 & 0.666 \\
5 & 17.87 & 18.04 & 18.07 & 0.234 & 0.222 & 0.305 & 0.447 & 0.485 & 0.640 \\
6 & 17.22 & 17.41 & 17.26 & 0.240 & 0.219 & 0.292 & 0.435 & 0.477 & 0.609 \\
7 & 16.86 & 17.02 & 16.62 & 0.238 & 0.217 & 0.308 & 0.410 & 0.434 & 0.593 \\
\midrule
\multicolumn{10}{l}{\emph{vs.\ AIFS ($n{=}66$)}} \\
1 & 21.02 & 21.61 & 21.48 & 0.376 & 0.319 & 0.376 & 0.726 & 0.719 & 0.755 \\
2 & 21.07 & 21.66 & 20.30 & 0.343 & 0.287 & 0.414 & 0.650 & 0.648 & 0.734 \\
3 & 20.58 & 21.10 & 19.55 & 0.357 & 0.305 & 0.430 & 0.587 & 0.594 & 0.718 \\
4 & 19.84 & 20.21 & 19.01 & 0.349 & 0.305 & 0.422 & 0.544 & 0.542 & 0.689 \\
5 & 19.11 & 19.25 & 18.35 & 0.332 & 0.308 & 0.409 & 0.503 & 0.529 & 0.679 \\
6 & 18.25 & 18.61 & 17.14 & 0.321 & 0.276 & 0.425 & 0.460 & 0.480 & 0.635 \\
7 & 18.28 & 18.52 & 16.63 & 0.294 & 0.261 & 0.458 & 0.421 & 0.418 & 0.618 \\
\midrule
\multicolumn{10}{l}{\emph{vs.\ GraphCast ($n{=}66$)}} \\
1 & 21.02 & 21.61 & 20.86 & 0.376 & 0.319 & 0.379 & 0.726 & 0.719 & 0.729 \\
2 & 21.07 & 21.66 & 20.24 & 0.343 & 0.287 & 0.416 & 0.650 & 0.648 & 0.694 \\
3 & 20.58 & 21.10 & 19.78 & 0.357 & 0.305 & 0.416 & 0.587 & 0.594 & 0.682 \\
4 & 19.84 & 20.21 & 19.14 & 0.349 & 0.305 & 0.412 & 0.544 & 0.542 & 0.656 \\
5 & 19.11 & 19.25 & 18.25 & 0.332 & 0.308 & 0.409 & 0.503 & 0.529 & 0.648 \\
6 & 18.25 & 18.61 & 16.95 & 0.321 & 0.276 & 0.435 & 0.460 & 0.480 & 0.602 \\
7 & 18.28 & 18.52 & 16.46 & 0.294 & 0.261 & 0.457 & 0.421 & 0.418 & 0.571 \\
\bottomrule
\end{tabular}%
}
\caption{Cross-model skill by lead (per-lead detail behind Table~\ref{tab:supp-crossmodel-summary}
above), restricted to the cells M-JEPA's tracked patches cover, the same patch-native frame as
main-paper Table~1 and main-paper Figure~3. RMSE (mm/day) $\downarrow$,
ACC $\uparrow$, FSS $\uparrow$ ($10$\,mm, $5^\circ$),
case-averaged. ``det.'' is the deterministic head; ``ens.'' is the probability branch's $16$ raw
members averaged into a point forecast (Table~\ref{tab:deterministic}'s ensemble-mean row). Window
count per lead: $161$
(IFS), $66$ (AIFS/GraphCast); pooled across leads these give the $266$/$127$ distinct target days of
Table~\ref{tab:supp-crossmodel-summary}. All three references lead on ACC and FSS at every lead. On RMSE, the
deterministic head is lower than IFS at leads~1--6 (IFS regains it only at lead~7) but never leads
AIFS or GraphCast; the ensemble mean's RMSE advantage over IFS is narrower still (leads~1 and~5 only) and
never leads AIFS/GraphCast.}
\label{tab:supp-crossmodel}
\end{table*}

\begin{figure}[h]
\centering
\includegraphics[width=0.85\linewidth]{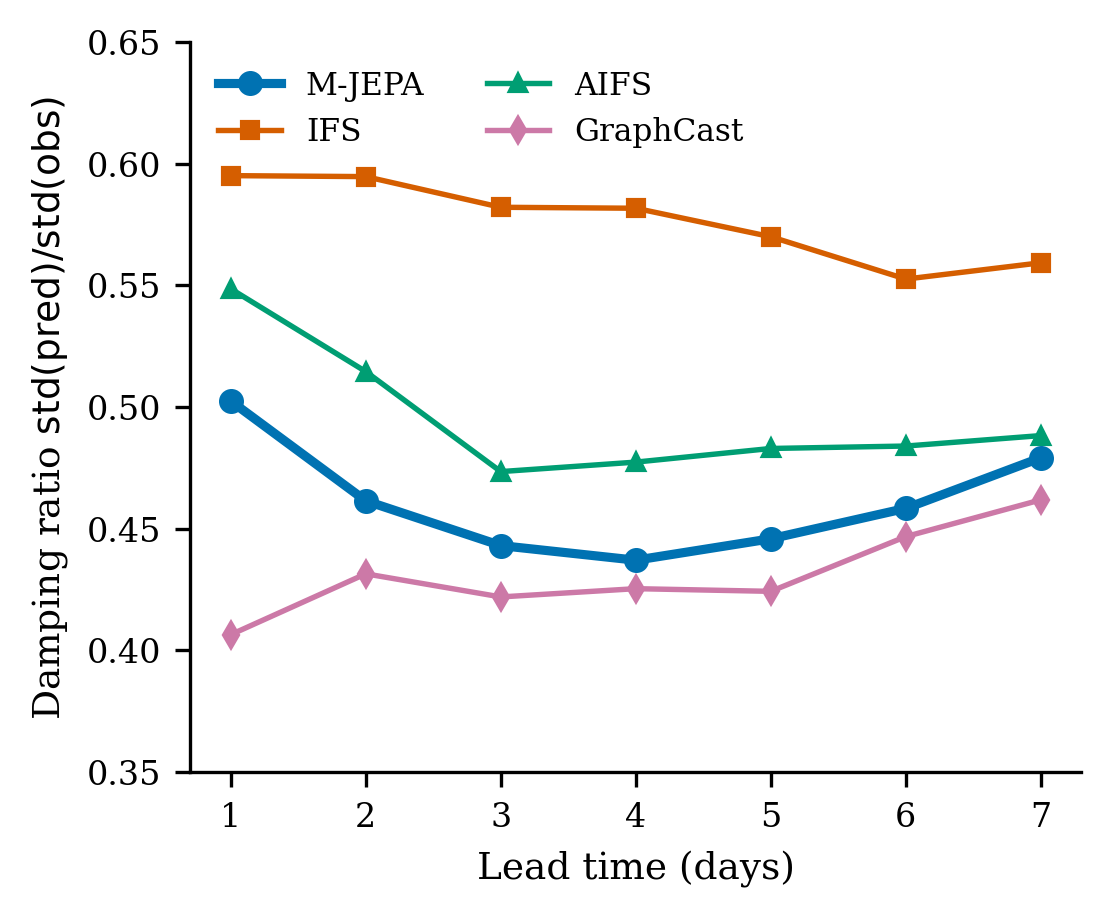}
\caption{Amplitude-stability comparison among actual forecasting systems (persistence excluded,
it relays a stale, systematically too-large field and runs to $1.31$ by lead~7, off this axis):
damping ratio $\mathrm{std}(\mathrm{pred})/\mathrm{std}(\mathrm{obs})$ by lead, patch frame,
case-averaged. IFS retains the most amplitude ($\approx0.55$--$0.60$, closest to $1$), M-JEPA and
AIFS track a comparable, non-collapsing band ($\approx0.44$--$0.55$), and GraphCast sits lowest
throughout ($\approx0.41$--$0.46$), consistent with its already-smooth-from-lead-1 signature. M-JEPA
shows no distinct advantage on this diagnostic; the point is that it does not exploit the
amplitude-collapse route to inflated ACC that AIFS's declining ratio suggests.}
\label{fig:supp-damping}
\end{figure}

\subsection{The Archives' Rising-ACC Pattern: Mechanism and a Full-Domain Check}
\label{sec:supp-acc-trend}

Table~\ref{tab:supp-crossmodel} shows AIFS and GraphCast's own ACC \emph{rising} and RMSE
\emph{falling} with lead (AIFS: ACC $0.376\to0.458$, RMSE $21.48\to16.63$; GraphCast: ACC
$0.379\to0.457$, RMSE $20.86\to16.46$), the physically backwards direction for a forecast skill
curve.

\paragraph{Mechanism.} Because ACC is scale-invariant (Section~\ref{sec:supp-verification}), a
forecast that loses fine-scale detail relative to the truth as lead grows can show \emph{rising} ACC
even as real skill falls: the residual correlation concentrates on the retained large-scale pattern once
high-frequency, largely uncorrelated detail is filtered out. We test this with two per-lead
diagnostics on the same patch population: a damping ratio ($\mathrm{std}(\mathrm{pred})/
\mathrm{std}(\mathrm{obs})$) and a roughness ratio (mean absolute nearest-neighbor gradient of
forecast over obs, a blur signature amplitude alone can miss); Figure~\ref{fig:supp-damping} shows
the damping ratio by lead for all four systems. AIFS's damping ratio falls
$0.549\to0.488$ and its roughness ratio falls $0.396\to0.323$ over the horizon: a genuine, progressive
self-smoothing that mechanically raises ACC under a scale-invariant metric. GraphCast's roughness
ratio is flat over the same horizon ($\approx0.27$--$0.29$): it is uniformly over-smooth from
lead~1 rather than increasingly so, while the observed field's own roughness falls $22\%$ ($9.41\to7.35$)
over the horizon, matching the event-lifecycle decay already established in
main-paper Section~5.3: as the tracked system ages, the truth itself simplifies toward
something a persistently smooth forecast pattern-matches more easily, without GraphCast itself
changing.

\begin{figure}[t]
\centering
\includegraphics[width=0.75\linewidth]{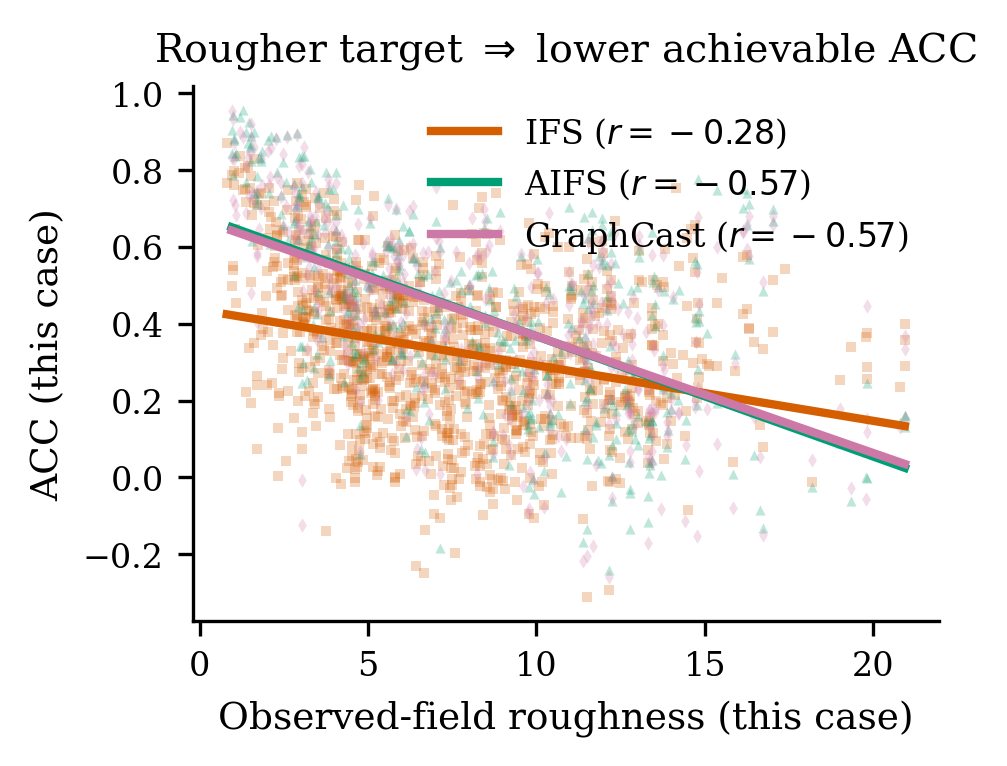}
\caption{The ACC-mechanism, tested directly: every (window, lead) case's ACC vs.\ that
case's own observed-field roughness, pooled over leads, with fitted trend and Pearson $r$ per archive.}
\label{fig:acc-mechanism}
\end{figure}

\paragraph{Case-level test of the mechanism.} The lead-aggregated numbers above are only 7 points per
archive and their ACC does not fall monotonically lead-to-lead, so we test the causal claim directly
and with far more statistical power: every individual (window, lead) case's ACC scattered against
that same case's own observed-field roughness, pooling across all leads
(Figure~\ref{fig:acc-mechanism}). The relationship is negative and substantial at case-level
resolution ($n{=}462$--$1127$ per archive): Pearson $r=-0.28$ (IFS), $-0.57$ (AIFS), $-0.57$
(GraphCast). Rougher, more complex target cases achieve lower ACC regardless of which lead they come
from, and smoother, simpler cases achieve higher ACC regardless of lead; lead enters only because it
correlates with case roughness (the lifecycle-decay effect), not because ACC directly rewards longer
lead.

\paragraph{Full-domain check.} Both patch-frame mechanisms above are consistent with the trend being
a property of the \emph{tracked, aging system} the patch follows, not of the archives' skill at long
lead in general. We test this directly: the same forecast files and target days, rescored on the full
$40\times40$ domain the reference systems natively forecast, instead of the patch crop around the
track (Table~\ref{tab:supp-fulldomain}). The rising-ACC / falling-RMSE pattern disappears entirely---
ACC is flat and near zero for all three systems across the full horizon, and RMSE mildly
\emph{increases} with lead, the physically expected direction. Full-domain ACC is also far lower in
absolute terms than the patch-frame values ($\lesssim0.04$ vs.\ $0.29$--$0.46$), as expected: the
patch frame's climatology reference captures structure specific to an active convective envelope,
while the full domain dilutes that signal with mostly quiescent surrounding area. The disappearance of
the trend rules out a genuine improvement in archive skill at longer lead; it is consistent with an
artifact of scoring one aging system's patch through its life, the same effect underlying the
observed amplitude decay reported in main-paper Section~5.3. Age and lead covary in this design and
are not separately identified.

\begin{table}[t]
\centering\small
\caption{Full-domain ($40\times40$) rescoring of the reference archives, case-averaged, on the
Table~\ref{tab:supp-crossmodel} target days for which a complete full-domain reference field exists;
that requirement drops roughly a tenth of the windows ($145/63/63$ against $161/66/66$), since a
patch-frame score needs only the patch cells whereas a full-domain score needs the whole grid. RMSE (mm/day) $\downarrow$, spatial anomaly
correlation $\uparrow$. Unlike the patch-frame comparison, no patch is centered on the tracked system
here, so any lifecycle-decay effect specific to the tracked system cannot appear.}
\label{tab:supp-fulldomain}
\resizebox{\linewidth}{!}{%
\begin{tabular}{lcccccc}
\toprule
& \multicolumn{2}{c}{IFS ($n{=}145$)} & \multicolumn{2}{c}{AIFS ($n{=}63$)} & \multicolumn{2}{c}{GraphCast ($n{=}63$)} \\
Lead & RMSE & ACC & RMSE & ACC & RMSE & ACC \\
\midrule
1 & 19.73 & $-0.020$ & 20.52 & 0.020 & 19.24 & 0.032 \\
2 & 19.78 & $-0.017$ & 20.14 & 0.029 & 19.19 & 0.040 \\
3 & 20.03 & $-0.026$ & 19.96 & 0.025 & 19.19 & 0.035 \\
4 & 20.26 & $-0.035$ & 19.95 & 0.007 & 19.36 & 0.023 \\
5 & 20.34 & $-0.030$ & 20.21 & 0.003 & 19.89 & 0.017 \\
6 & 20.69 & $-0.023$ & 20.35 & 0.010 & 20.26 & 0.022 \\
7 & 20.74 & $-0.020$ & 20.32 & 0.013 & 20.34 & 0.023 \\
\bottomrule
\end{tabular}%
}
\end{table}

\subsection{Ensemble Mean by Lead}
\label{sec:supp-ensmean}

Table~\ref{tab:supp-ensmean-persist} gives the per-lead detail for
Table~\ref{tab:deterministic}'s ensemble-mean row (the probability branch's
$16$ raw members averaged into a point forecast), same conventions and populations as
Tables~\ref{tab:supp-deterministic}--\ref{tab:supp-crossmodel}, true patch frame throughout.

\begin{table}[h]
\centering\small
\begin{tabular}{lcc}
\toprule
Lead (days) & Ensemble mean & Persistence \\
\midrule
\multicolumn{3}{l}{\emph{RMSE (mm/day) $\downarrow$}} \\
1 & 17.20 & 21.12 \\
2 & 17.46 & 23.97 \\
3 & 17.12 & 24.30 \\
4 & 16.62 & 23.96 \\
5 & 16.13 & 23.73 \\
6 & 15.68 & 23.69 \\
7 & 15.13 & 23.45 \\
\midrule
\multicolumn{3}{l}{\emph{Spatial ACC $\uparrow$}} \\
1 & 0.409 & 0.384 \\
2 & 0.328 & 0.201 \\
3 & 0.291 & 0.148 \\
4 & 0.259 & 0.139 \\
5 & 0.256 & 0.129 \\
6 & 0.257 & 0.122 \\
7 & 0.244 & 0.114 \\
\midrule
\multicolumn{3}{l}{\emph{FSS10 at $5^\circ$ neighborhood $\uparrow$}} \\
1 & 0.726 & 0.756 \\
2 & 0.641 & 0.606 \\
3 & 0.573 & 0.548 \\
4 & 0.513 & 0.536 \\
5 & 0.486 & 0.513 \\
6 & 0.476 & 0.493 \\
7 & 0.450 & 0.473 \\
\bottomrule
\end{tabular}
\caption{Ensemble mean vs.\ persistence by lead, full test set ($n{=}337$), case-averaged. Unlike
the pooled convention, the ensemble mean's ACC now exceeds persistence's at every lead, including
lead~1. FSS10 favors the ensemble mean at leads~2--3 only and persistence at leads~1 and~4--7, the
same crossover pattern as the deterministic branch (Table~\ref{tab:supp-deterministic}).}
\label{tab:supp-ensmean-persist}
\end{table}

The ensemble mean's cross-model detail (vs.\ IFS/AIFS/GraphCast) is merged into
Table~\ref{tab:supp-crossmodel} above as the ``ens.'' columns alongside the deterministic head.

\subsection{Extreme-Event Skill of the Probability Branch}
\label{sec:supp-extreme}

Grid-point categorical scores (POD: probability of detection; FAR: false alarm ratio;
CSI: critical success index; ETS: equitable threat score) are strict single-cell measures with no
neighborhood tolerance, so they isolate exact-hit skill from the placement credit FSS gives nearby
cells. On the matched $n{=}169$ population in the true patch frame, member-averaged POD at Heavy is
$0.053$ against the operational ensemble's $0.041$, and at Very Heavy $0.023$ against $0.015$; the
ordering reverses only at Extremely Heavy ($0.006$ vs.\ $0.013$, $230$ matched fields, the thinnest
sample of the three tiers).

Member-FSS at that third tier behaves the same way and is the one threshold where the main paper's
extreme-skill result does not hold. The ordering is lead-dependent: IFS Ens leads at the first two
leads ($0.069$ against $0.196$ at lead~1; $0.062$ against $0.144$ at lead~2, M-JEPA first in each
pair), M-JEPA Ens leads from lead~3 onward and by up to $6.6\times$ at lead~5 ($0.046$ against
$0.007$), and pooled over leads the short-lead deficit dominates ($0.056$ against $0.077$, $230$
matched fields). With $27$--$38$ qualifying events per lead this tier is too thin to support either direction.

\subsection{Directional Tracking and Rollout Velocity of the Decoded Forecast}
\label{sec:supp-propagation}

\begin{figure}[t]
\centering
\includegraphics[width=0.85\linewidth]{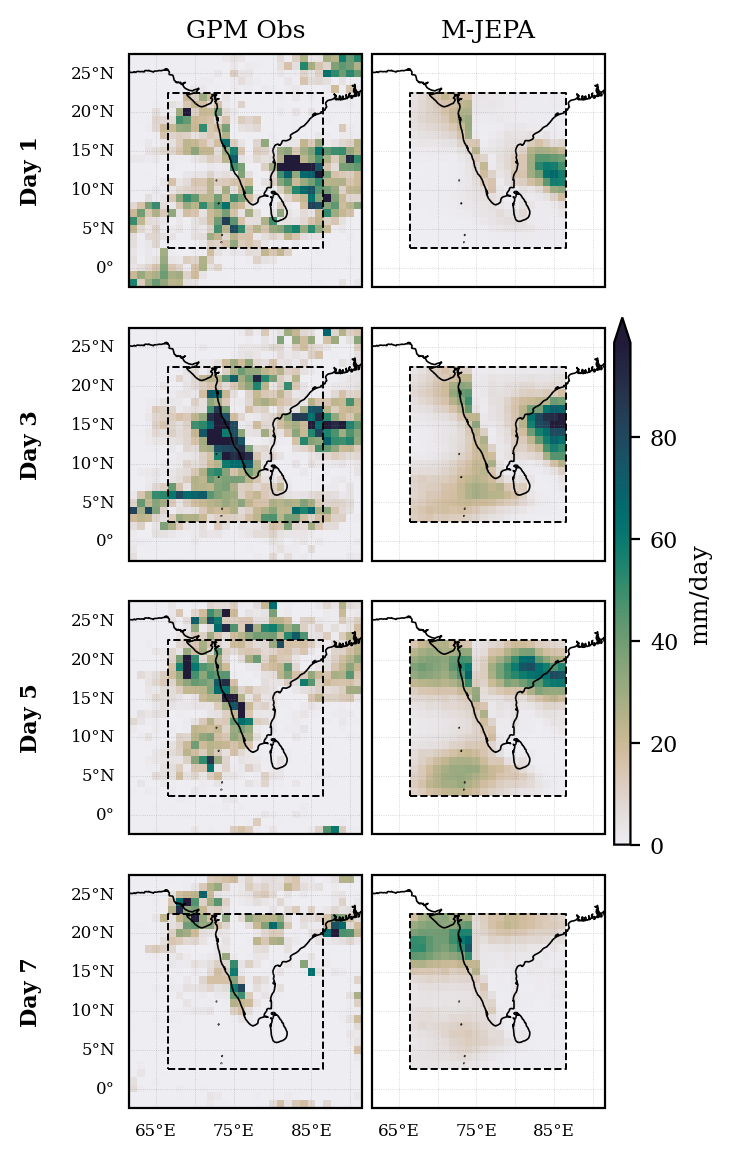}
\caption{GPM observations (left) and the M-JEPA deterministic field (right) at days $1,3,5,7$ of the
seven-day forecast for the wettest test window by observed day-3 rainfall (target days 2023-07-02 to
2023-07-08). The forecast places a maximum on the observed coastal cell at short range and broadens
into a smoother band by day~7. Dashed box marks the native patch extent.}
\label{fig:lifecycle-case}
\end{figure}

\begin{figure*}[t]
\centering
\includegraphics[width=\textwidth]{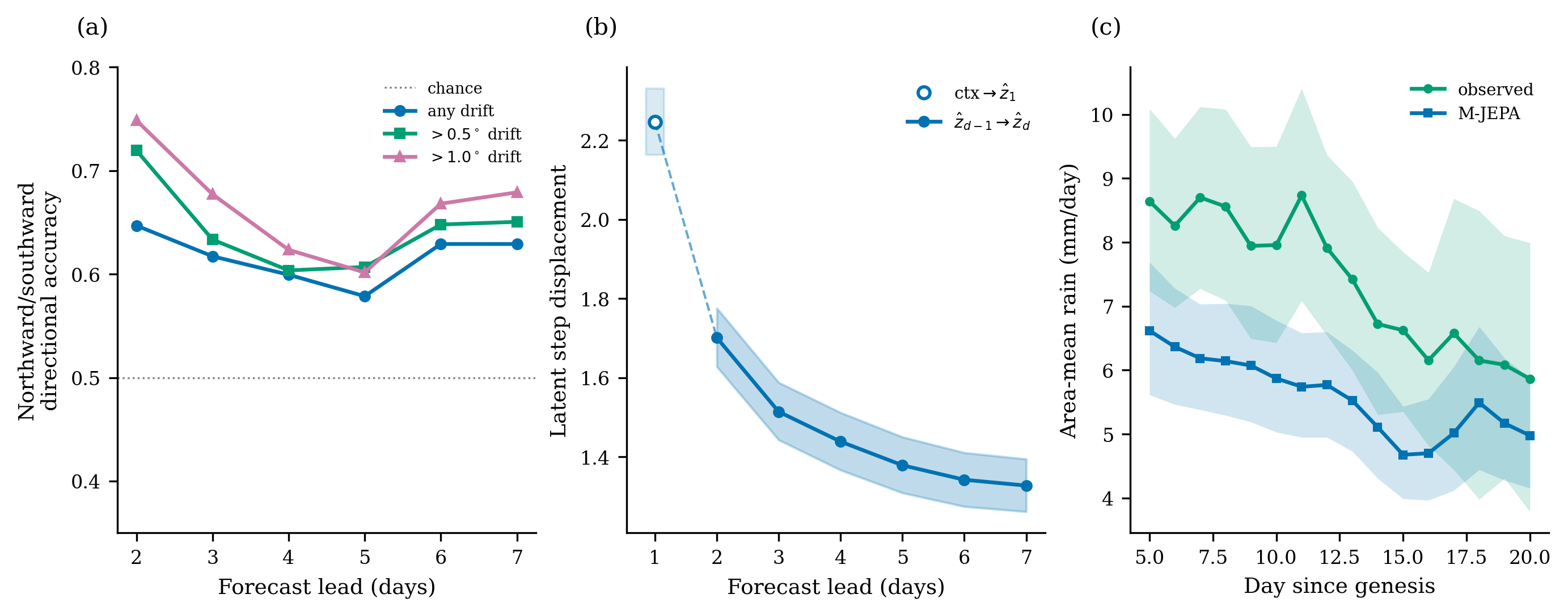}
\caption{Pooled evolution diagnostics over the full test set ($n{=}337$), rain-centroid based
($1$ cell $=1^\circ$; $95\%$ window-bootstrap bands where applicable). (\textbf{a}) Northward/southward
directional accuracy of the decoded centroid's drift relative to lead~1 (sign match against observed),
at three observed-drift thresholds: above the $50\%$ chance line at every lead and every threshold,
and rising as the threshold tightens (near-zero-displacement noise excluded). (\textbf{b})
Step-to-step latent rollout displacement $\|\hat z_d - \hat z_{d-1}\|_2$ against lead (pooled
foreground latent; the open marker and dashed segment mark the context$\to\hat z_1$ step, a different
regime from the subsequent $\hat z_{d-1}\to\hat z_d$ steps). Displacement decays from $2.25$ to a
non-zero floor of $1.33$--$1.38$ by leads~5--7 ($\approx 1.69\times$ damping, early against late
step): the latent rollout is damped but not frozen. (\textbf{c}) Composite intensity lifecycle:
observed and decoded area-mean rain by day since genesis, pooled over $49$ test tracks
(window-bootstrap bands; days with fewer than $20$ pooled samples omitted). Normalized to each
series' own day-5 value, the decay rates over days 5--16 are $-0.0247$ and $-0.0253$ per day on a dry
bias of $2.1$\,mm/day; in the thinner tail of the longest-lived tracks (day~$17$ onward) the decoded
curve stops declining while the observed curve continues to fall. Panels (b) and (c) are two
signatures of the same rollout damping, on undecoded latent velocity and decoded amplitude
respectively; panel (a) is a directional result and does not speak to damping.}
\label{fig:supp-propagation}
\end{figure*}

Figure~\ref{fig:lifecycle-case} shows one decoded forecast against observations across the horizon,
as a qualitative reference for the pooled diagnostics that follow.
Grid-point and neighborhood scores measure where rain is placed, not whether the forecast follows the
\emph{motion} of the convective envelope. To test motion at the decoded-forecast level, we take the
precip-weighted centroid (cells $>1$\,mm/day) of the observed and forecast rain field at each lead,
for every held-out window ($n{=}337$), and evaluate a directional rather than a magnitude criterion:
whether the forecast centroid's drift relative to lead~1 has the same \emph{sign} (northward vs.\
southward) as the observed drift, regardless of how far it moves (Figure~\ref{fig:supp-propagation}a).
At the loosest threshold (any nonzero observed drift), directional accuracy is $58$--$65\%$ across leads,
against a $50\%$ chance rate ($n{=}2022$ pooled over leads~2--7, binomial $p<10^{-4}$); restricting to
observed drifts $>0.5^\circ$ raises this to $61$--$69\%$, and $>1.0^\circ$ to $63$--$71\%$ ($n{=}1355$
pooled): the signal strengthens, not weakens, as near-zero-displacement noise is excluded, the
signature of a genuine effect rather than an artifact of small splits. The same pattern holds on the
zonal (east/west) axis: pooled accuracy $59.0\%$ ($n{=}2022$, $p<10^{-4}$), rising to $64.0\%$
restricted to drifts $>1.0^\circ$. This is a directional result only, and says nothing about
displacement \emph{magnitude}, which the model does not resolve well (see below).

Section~\ref{sec:supp-q3}'s CKA plateau is a representational-structure diagnostic and does not by
itself establish whether the latent is still \emph{moving} once CKA saturates.
Figure~\ref{fig:supp-propagation}(b) answers that directly, with the raw step-to-step displacement of
the rollout in latent space, $\|\hat z_d - \hat z_{d-1}\|_2$ (pooled foreground latent, same pooling
as the linear-probe features), at every autoregressive step. Displacement falls from $2.25$ (context
$\to \hat z_1$) to $1.70$ (step into lead~2) and continues decaying toward a non-zero floor of
$1.33$--$1.38$ over leads~5--7, where the bootstrap CIs overlap and the curve is flat and no longer
shrinking. The ratio of the first step to the last (context$\to\hat z_1$ against lead~6$\to$7) is
$\approx 1.69\times$: rollout velocity falls to about $60\%$ of its initial rate but does not vanish. This refines the
rollout-saturation picture and does not contradict it: the CKA plateau shows the representation's
\emph{structure} stops changing, while panel~(b) shows the underlying latent state itself keeps
moving at a damped, non-zero rate, so saturation asymptotes to a nonzero floor rather than collapsing
the rollout to a fixed point.

Panel~(c) tests amplitude over the event's own lifecycle, which is the axis on which the forecast
does track the observed evolution. For every held-out window we compute the day since the track's
genesis for each of its seven forecast-lead target days, and pool the observed and decoded area-mean
rain intensity by that lifecycle day across the $49$ distinct tracks that supply the $337$ evaluation
windows (a subset of the $149$ detected test-year tracks in Figure~\ref{fig:supp-all-tracks}: a track
enters the evaluation set only where it yields a valid context-plus-horizon window at an unmasked
anchor, and each contributing track supplies several overlapping windows). Context length fixes a
floor of five days before any target day is reachable, so the earliest lifecycle day observable by
this construction is day~$5$, a coverage limit of the windowing, not a selection choice. From
day~$5$ through day~$16$ the decoded curve's \emph{shape} tracks the observed decay closely: each
series normalized to its own day-$5$ value falls to $0.71\times$ (observed) and $0.71\times$
(forecast) by day~$16$, decay rates of $-0.0247$ and $-0.0253$ per day, on a dry bias of
$2.1$\,mm/day over those days. Beyond day~$16$, in the thinner-sampled tail of longest-lived tracks
($n{=}84$ falling to $n{=}42$ by day~$20$), the observed curve continues falling to $0.68\times$
while the decoded curve stops declining at $0.75\times$: the forecast fails to continue the observed
decay for the longest-lived systems, which is the same damping seen in the CKA plateau and the latent
displacement floor, expressed in decoded amplitude at increasing lifecycle duration. Because the day
$5$--$16$ and post-day-$16$ ranges behave differently, both are reported; the split is not a
sampling threshold, since day~$17$ ($n{=}84$) is better sampled than day~$5$ ($n{=}40$).

\subsection{Radial Power Spectra of the Ensemble Members}
\label{sec:supp-spectra}

Main-paper Figure~2b compares the spatial power spectrum of each ensemble's members against
observations, following the azimuthally-averaged spectral-realism check used for generative weather
models \citep{gencast}. Because a Lagrangian patch is a $20\times20$ window rather than a global
field, we use a 2-D radial spectrum in place of a spherical-harmonic one. Each field is
mean-subtracted and multiplied by a 2-D Hann window to suppress edge leakage, transformed with
\texttt{fft2}, and its squared magnitude binned by integer radius $r=\lfloor\sqrt{(i-c_y)^2+(j-c_x)^2}\rfloor$
about the shifted spectral center; the $r{=}0$ (DC) bin is discarded, leaving radii $1$--$9$. Each
system's curve is the ratio of its mean binned power to the observed mean binned power in the same
bin, so a value of $1$ means the forecast carries exactly the observed variance at that scale.

Averaging runs over every member, every lead and every window of the matched population: $1183$
observed fields ($169$ windows $\times$ $7$ leads), $18\,928$ M-JEPA member fields ($\times16$) and
$60\,333$ IFS Ens member fields ($\times51$). Both systems lose power at every radius, so the
informative quantity is the \emph{direction} of the loss with wavenumber, not its level. IFS Ens
declines monotonically ($0.484$ at $r{=}1$ to $0.170$ at $r{=}9$), the expected signature of a
mass-conserving smoother, which spreads a sharp feature over a wider area at lower amplitude.
M-JEPA Ens begins slightly lower ($0.464$) and rises to $0.722$ at $r{=}8$ and $0.711$ at $r{=}9$,
$4.2\times$ IFS Ens's retention there. This is a descriptive comparison over pooled fields, not a
hypothesis test, and we attach no confidence interval to the ratios. It also speaks only to the
amount of small-scale variance, not to its placement, which is why the main text reads it together
with the member-FSS result.

\section{Scaling Hypothesis: Information-Limited Signal}
\label{sec:supp-scaling}

The rain-only control's pretraining window count is $226$ against $3885$ for the all-season
five-proxy pretraining signal, a $17\times$ reduction in information. Because the control is trained
on rainfall alone, this same rain-valid window population serves both its pretraining and its decoder
training, whereas M-JEPA pretrains on the $3885$ all-season proxy windows and trains its decoder on
the $667$ rain-labeled subset. Despite this, the control's skill degrades gracefully instead of
collapsing (CRPS $7.52$ against $5.54$\,mm/day, with Brier skill closest to zero among the rain-only
models; main-paper Table~3). A collapse-scale degradation would point to an idea-limited method, in
which the architecture or loss needs the specific target signal to work at all; the graceful
degradation observed instead is consistent with an information-limited method, where performance
tracks the richness of the pretraining signal rather than a narrow recipe requirement. We advance
this as a hypothesis supported by the control comparison, not as a
measured scaling law across multiple window-count settings; establishing a full scaling curve
(pretraining skill as a function of proxy window count, proxy channel count, and temporal
resolution) is the natural next step.

Temporal resolution is the other information axis. All three deterministic references
that outperform M-JEPA on point-forecast metrics (main-paper Section~5.2) resolve atmospheric state well below
the daily cadence of our pipeline: IFS's dynamical core integrates continuously, and both AIFS and
GraphCast step autoregressively at 6-hourly intervals \citep{aifs,graphcast}. M-JEPA's five proxies,
by contrast, are resampled to daily means before the encoder ever sees them, and the rollout itself
advances one calendar day per autoregressive step; any sub-daily structure in the underlying fields,
including the diurnal cycle of convective initiation and the nocturnal propagation of mesoscale
convective systems characteristic of the Bay of Bengal, is discarded at the data-ingestion stage, not
merely under-modeled. This is consistent with denser temporal sampling being one of the
information-ceiling levers named above, though the three references also differ from M-JEPA in
initial-condition quality and training-corpus scale, so temporal resolution cannot be isolated as the
cause from this comparison alone; a controlled ablation at matched initialization and data volume,
varying only proxy cadence, is the direct test.

\section{Computational Cost and Runtime}
\label{sec:supp-compute}

Model size: encoder and autoregressive predictor $9.7$M parameters combined
($4.8$M and $4.9$M respectively), decoder $5.7$M parameters (frozen Stage-1 stack, trainable
decoder heads only at readout time), for the $15.4$M total quoted in the main text. Inference cost splits by head. The deterministic point-forecast
head is a single forward pass: a full seven-day forecast for a batch of $16$ patches runs in
$204$\,ms on a single GPU ($12.8$\,ms per patch, amortized), with no iterative sampling.
The ensemble-generating flow probability head used for the matched IFS-Ens comparison (``Forecast Skill''
section, main paper) instead requires $30$ rectified-flow integration steps per ensemble member, and is
substantially more expensive per forecast than the deterministic head, though it still runs on the
same single GPU; members are scored as generated, with no post-hoc
recalibration step. No comparable figure exists for the operational ensemble in a form we can
measure, so the main text makes no quantitative compute comparison against IFS. We did not use the flow head for the probing
suite (Section~\ref{sec:supp-probing}), which reads out from the frozen encoder alone
and does not invoke either decoder head.

Pretraining hardware and software. All training and inference for this work ran on a single NVIDIA
RTX A4000 (16\,GB) under Ubuntu 22.04.5 LTS, with Python 3.13.1, PyTorch 2.6.0 (CUDA 12.4,
cuDNN 9.1.0), NumPy 2.2.2, SciPy 1.15.1, scikit-learn 1.6.1, xarray 2025.1.2, and pandas 2.3.3.
Every run seeds Python, NumPy, and PyTorch (CPU and CUDA) from a single configured value
(default $1410$) and disables cuDNN autotuning in favor of deterministic kernels; the
seed-sensitivity figures reported above use seeds $1410$, $42$, and $7$.
The self-supervised encoder pretraining recipe was: VICReg anti-collapse objective,
auxiliary OLR-reconstruction weight $0.10$, batch size $32$, and $100$ epochs; Stage~1 uses no
spatial jitter, which is applied only when training the Stage-2 decoder
($\pm2$ grid cells, $\approx\pm2^\circ$). The VICReg coefficients were variance
$1.0$, covariance $0.04$, and $\varepsilon = 10^{-4}$. Optimization used AdamW with
learning rate $5\times10^{-4}$, weight decay $0.05$, and a $10$-epoch warmup.

\end{document}